%% file: main.tex
\documentclass[sigconf]{acmart}
\usepackage[ruled]{algorithm2e}
\usepackage{amsmath,amsfonts}
\usepackage{tikz}
\usepackage{pseudo}
\usepackage{subfigure}
\usepackage{mathrsfs} 
\usepackage{multicol}
\usepackage{multirow}
\usepackage{bbding}
\usepackage{bm}
\usepackage{enumitem}
\usepackage{verbatim}
\usepackage{graphicx}
\usepackage{xspace}
\usepackage[normalem]{ulem}
\useunder{\uline}{\ul}{}
\usepackage{appendix}

\usepackage{enumitem}[nosep]
\setlist[itemize]{noitemsep, topsep=0pt,partopsep=0pt}

\AtBeginDocument{%
  \providecommand\BibTeX{{%
    \normalfont B\kern-0.5em{\scshape i\kern-0.25em b}\kern-0.8em\TeX}}}

\copyrightyear{2023}
\acmYear{2023}
\setcopyright{acmcopyright}
\acmConference[WSDM '23] {Proceedings of the Sixteenth ACM International Conference on Web Search and Data Mining}{February 27-March 3, 2023}{Singapore, Singapore.}
\acmBooktitle{Proceedings of the Sixteenth ACM International Conference on Web Search and Data Mining (WSDM '23), February 27-March 3, 2023, Singapore, Singapore}
\acmPrice{15.00}
\acmISBN{978-1-4503-9407-9/23/02}
\acmDOI{10.1145/3539597.3570369}

\newcommand{\eat}[1]{}
\newcommand{\setcmd}[1]{\mathcal{#1}}
\newcommand{\veccmd}[1]{\mathbf{#1}}
\newcommand{\matcmd}[1]{\mathbf{#1}}
\newcommand{\stitle}[1]{\vspace{0.5ex}{\bf #1}}  

\newcommand{\model}{{RTGNN}\xspace}

\begin{document}

\title{Robust Training of Graph Neural Networks via  Noise Governance}

\author{Siyi Qian}
\affiliation{%
  \institution{Zhejiang University}
  \city{Hangzhou}
  \country{China}
}
\email{syqian99@zju.edu.cn}

\author{Haochao Ying}
\authornote{Corresponding Author.}
\affiliation{%
  \institution{Zhejiang University}
   \city{Hangzhou}
  \country{China}
}
\email{haochaoying@zju.edu.cn}

\author{Renjun Hu}
\affiliation{%
  \institution{Alibaba Group}
   \city{Hangzhou}
  \country{China}
}
\email{renjun.hrj@alibaba-inc.com}

\author{Jingbo Zhou}
\affiliation{%
  \institution{Baidu Research}
   \city{Beijing}
  \country{China}
}
\email{zhoujingbo@outlook.com}

\author{Jintai Chen}
\affiliation{%
  \institution{Zhejiang University}
   \city{Hangzhou}
  \country{China}
}
\email{jtchen721@gmail.com}

\author{Danny Z. Chen}
\affiliation{%
  \institution{University of Notre Dame}
    \city{Notre Dame}
  \country{USA}}

\email{dchen@nd.edu}

\author{Jian Wu}\affiliation{%
  \institution{Zhejiang University}
   \city{Hangzhou}
  \country{China}
  }\email{wujian2000@zju.edu.cn}

\renewcommand{\shortauthors}{Siyi Qian, et al.}


\input{files/0.abstract}

\ccsdesc[500]{Information systems~Data mining}

\keywords{GNNs; semi-supervised learning; label noise; robustness}


\maketitle
\input{files/1.introduction}

\input{files/2.related_works}

\input{files/3.preliminaries}

\input{files/4.methodology}

\input{files/5.experiment}

\input{files/6.conclusion}
\section*{Acknowledgements}
This research was partially supported by National Key R\&D Program of China under grant No. 2018AAA0102102, National Natural Science Foundation of China under grants No. 62132017 and No. 62106218.
\hfill
\bibliographystyle{ACM-Reference-Format}
\bibliography{main}

\end{document}

%% file: files/0.abstract.tex
\begin{abstract}
Graph Neural Networks (GNNs) have become widely-used models for semi-supervised learning. However, 
the robustness of GNNs in the presence of label noise remains a largely under-explored problem. In this paper, we consider an important yet challenging scenario where labels on nodes of graphs are not only noisy but also scarce. 
In this scenario, the performance of GNNs is prone to degrade due to label noise propagation and insufficient learning. To address these issues, we propose a novel \model (\underline{R}obust \underline{T}raining of \underline{G}raph \underline{N}eural \underline{N}etworks via Noise Governance) framework that achieves better robustness by learning to explicitly govern label noise.
More specifically, we introduce self-reinforcement and consistency regularization as supplemental supervision. The self-reinforcement supervision is inspired by the memorization effects of deep neural networks and aims to correct noisy labels. Further, the consistency regularization prevents GNNs from overfitting to noisy labels via mimicry loss in both the inter-view and intra-view perspectives.
To leverage such supervisions, we divide labels into clean and noisy types, rectify inaccurate labels, and further generate pseudo-labels on unlabeled nodes. Supervision for nodes with different types of labels is then chosen adaptively. This enables sufficient learning from clean labels while limiting the impact of noisy ones. 
%
We conduct extensive experiments to evaluate the effectiveness of our \model framework, and the results validate its consistent superior performance over state-of-the-art methods with two types of label noises and various noise rates.
\end{abstract}

%% file: files/1.introduction.tex
\section{Introduction} \label{sec:intro}

In real-world applications, a set of objects and their relationships can often be represented naturally as a graph. The graph data structure is widely employed in various domains such as biology, transportation, and social science. In the past decade, Graph Neural Networks~(GNNs) have shown promising 
capacity
in modeling graph data~\cite{kipf2016semi,hamilton2017inductive,xu2018powerful}. Typically, GNNs adopt a message passing and aggregation procedure to effectively propagate information via the graph structure. This mechanism makes GNNs very suitable for semi-supervised graph learning (e.g., node classification~\cite{alam2018graph,wang2019semi}). 



While GNNs are generally effective, most of the existing approaches assume that labels are sufficient and clean. However, in practice, 
node labels can be both scarce and noisy. 
For example, consider a graph from social media whose node labels are contributed by users. It is often the case that only a small fraction of users would participate in label generation, and intentionally or for other reasons, some of the labels do not reflect the truth. 
Another example is crowd-sourcing node labels, e.g., fake news annotation and medical knowledge graph annotation. It is easy to see that the annotation process is labor-intensive and expensive, and almost inevitably, label errors are introduced due to subjective judgment.
%
%
Such cases will lead to graphs with scarce and noisy node labels. It has already been observed that noisy labels could pose a severe threat to the generalization performance 
of deep learning models~\cite{arpit2017closer,zhang17understanding}.
Therefore, developing label-efficient and noise-resistant GNNs is 
an important and challenging problem.


In the literature, robust deep learning in the presence of noisy labels has been explored mainly in applications with non-graph data such as images. Several strategies have been developed to combat label noises, e.g., sample selection~\cite{huang2019o2u,han2018co,malach2017decoupling}, robust loss functions~\cite{ghosh2017robust,zhang2018generalized,wang2019symmetric}, and loss correction~\cite{patrini2017making,nt2019learning}.
However, simply incorporating these approaches into GNNs will be insufficient (we will empirically show this in experiments). We observe that one unique characteristic of this problem with GNNs is that the scarcity of labels causes difficulties for nodes to receive sufficient supervision from labeled neighbors. Meanwhile, nodes in graphs are directly influenced by potential noises through message passing.
Failing to balance the two would either lead to massive erroneous supervision from noisy labels or end up with insufficient learning.

Hence, a main challenge to solving this problem is how to \emph{effectively leverage supervision of clean labels while limiting the impact of noisy ones}. 
Previous studies mainly focused on leveraging supervision of labels, but neglected how to limit the impact of noisy labels.  
For example, recently, NRGNN~\cite{dai2021nrgnn} investigated a robust GNN with noisy and sparse labels. It proposed to link unlabeled nodes with labeled nodes and further mine accurate pseudo-labels to provide more supervision. However, NRGNN mainly emphasized leveraging supervision of labels, which mixed up clean and noisy labels in its learning process. Whereas, 
as explained above, explicitly governing noises~(i.e., limiting the impact of noisy labels) is necessary in order to further boost the robustness of GNNs.


To this end, we propose a novel GNN framework called \model (\underline{R}obust \underline{T}raining of \underline{G}raph \underline{N}eural \underline{N}etworks via Noise Governance) 
that is capable of conducting robust learning with scarce and noisy node labels. 
As a distinguished design of our model, we develop a fine-grained noise governance strategy in semi-supervised learning. Due to the fact that deep neural networks~(DNNs) tend to prioritize learning simple patterns first and may overfit to noises~\cite{arpit2017closer}, we first propose an effective and scalable loss-based label division method to identify potentially noisy labels. Further, besides the commonly-used label supervision, we introduce self-reinforcement and consistency regularization as supplemental supervisions. The self-reinforcement supervision is inspired by the memorization effects of DNNs and aims to correct noisy labels.
Consistency regularization includes an inter-view regularization based on ensembled classifiers capable of filtering different errors, and an intra-view regularization to explore the local homogeneity of the graph~\cite{yang2016revisiting}.

Specifically, we propose to train a pair of peer GNNs to enforce adequate supervision as well as govern noise labels. First, we augment the raw graph by linking labeled and unlabeled nodes to facilitate efficient message passing, following NRGNN~\cite{dai2021nrgnn}. Then, in each epoch, we progressively perform the main division of clean and noisy candidate set by the small-loss criterion~\cite{han2018co}. Next, a subset of confident nodes which predict a different class from their labels in the noisy candidate set is reinforced by training with their own prediction. Similarly, we extend the label set by adding those confident and consistent unlabeled nodes to the training set. Finally, we apply inter-view regularization to help two classifiers cooperatively mimic each other's soft targets and intra-view regularization to enable nodes to learn from their neighbors. 

In summary, the main contributions of our work are as follows.
\begin{itemize}
\item We investigate the robust training problem of GNNs from the noise governance perspective, which has been under-explored in previous studies.
\item We develop a novel \model model which governs label noises explicitly with self-reinforcement and consistency regularization. \model also enables fine-grained learning on relatively clean, potentially noisy, and pseudo labels.
This allows effectively leveraging supervision information while limiting the impact of label noises.
\item We conduct extensive experiments to evaluate the effectiveness of our new approach, and the results validate the consistent superior performance of \model over state-of-the-art methods with two types of noises and various noise rates.
\end{itemize}

%% file: files/2.related_works.tex
\section{Related Work}
\label{sec:relatedWork}


\subsection{Robust Deep Learning}
\label{subsec:robust}

It has been reported in~\cite{han15deep,louizos18learning} that DNNs are prone to over-parameterize. As a result, they could 
overfit to and even memorize noisy labels, resulting in degraded generalization and poor performance~\cite{arpit2017closer,zhang17understanding}. In the last few years, robust deep learning dealing with label noises is gaining more attention, especially in applications with non-graph data such as images~\cite{malach2017decoupling,jiang2018mentornet,han2018co,ghosh2017robust,zhang2018generalized,patrini2017making}. Such work can be roughly divided into two categories: sample-centric and loss-centric.

Sample-centric approaches seek to enhance the robustness of deep models by selecting clean samples for training. Malach \textit{et al}.~\cite{malach2017decoupling} proposed an update-by-disagreement strategy for sample selection, and trained two predictors simultaneously and performed parameter updates only on samples in disagreement areas of the two predictors. Inspired by the memorization effects of DNNs, Han \textit{et al}.~\cite{han2018co} developed a Co-teaching paradigm that also trains two networks such that each network samples its small-loss instances to update the parameters of its peer network. The small-loss Co-teaching paradigm has been further extended by bridging the update-by-disagreement strategy as Co-teaching+~\cite{yu2019does} and the Co-regularization max-agreement principle as JoCoR~\cite{wei2020combating}. Recently, more metrics have been explored for sample selection, e.g., loss at different stages of model fitness~\cite{huang2019o2u}, network output at different training epochs~\cite{nguyen2019self}, margins to decision boundaries~\cite{pleiss2020identifying}, sample uncertainty~\cite{xia2021sample}, and divergence between different classifiers~\cite{yu2021divergence}.

In contrast, loss-centric approaches adopt noise-robust loss functions or perform direct loss correction to combat label noises. One type of such methods aims to design loss functions that are robust to noise. Ghosh \textit{et al}.~\cite{ghosh2017robust} showed that the mean absolute error (MAE) loss is inherently robust to label noises. This idea was further extended by GCE~\cite{zhang2018generalized} which can be viewed as a generalization of MAE and cross entropy (CE). SCE~\cite{wang2019symmetric} boosts CE symmetrically with a noise-robust term called reverse cross entropy (RCE). The complementary loss function method~\cite{wang2021learning} utilizes both CE loss and robust loss to balance learning sufficiency and robustness. APL~\cite{ma2020normalized} classifies existing loss functions into active and passive based on their optimization behavior; it leverages active normalized loss for robust training and complements training with passive loss to address the under-fitting problem. The second type of methods is direct loss correction.  Goldberger \textit{et al.}~\cite{DBLP:conf/iclr/GoldbergerB17} adopted a noise adaptation layer to model and correct the noise distribution.  The F-correction method~\cite{patrini2017making} proposes forward and backward correction which rectifies loss based on the noise transition matrix.

While the above methods achieved good performance in supervised training on non-graph data, our work explores a new perspective that focuses on semi-supervised learning with noisy labels on graphs and designs a new noise governance strategy to alleviate noise impact.

\subsection{Training GNNs with Noisy Labels}
\label{sec:relatedwork:gnn}
GNNs have achieved state-of-the-art performance on several semi-supervised benchmarking tasks, e.g., link prediction and node classification~\cite{kipf2016semi,hamilton2017inductive,velivckovic2017graph,xu2018powerful}. GNNs broaden the convolution operation to graph data. They learn a graph node's representation by aggregating the features of its neighbors. GNNs can mainly be divided into spectral-based~\cite{bruna13spectral,defferrard2016convolutional,kipf2016semi} and spatial-based~\cite{velivckovic2017graph,hamilton2017inductive,xu2018powerful} ones.
While it has been shown that deep models are vulnerable to label noises, robust training of GNNs is generally under-explored, and only a few studies investigated the robustness of GNNs with noisy labels~\cite{nt2019learning,dai2021nrgnn}. 
Among them, D-GNN~\cite{nt2019learning} shows that GCNs could be very vulnerable to label noises, and exploits backward loss correction~\cite{patrini2017making} to boost performance. However, one-step loss correction is insufficient to deal with challenging scenarios, e.g., the one with noisy and scarce labels that we consider.
Close to our work is NRGNN~\cite{dai2021nrgnn}, which learns a robust GNN with noisy and sparse labels; it proposes to link unlabeled nodes with labeled ones and further mine accurate pseudo-labels to provide more supervision. However, noisy labels are not explicitly governed by NRGNN, thus resulting in error propagation during message passing to some extent. Different from NRGNN, we distinguish noisy labels from clean ones and apply additional supervision and label correction to reduce the impact of label noises.






%% file: files/3.preliminaries.tex
\section{Preliminaries}
\label{sec:preliminaries}


\stitle{Notation and problem formulation}.
We represent a graph $\mathcal{G}=(\setcmd{V}, \matcmd{A}, \matcmd{X})$ as a triple: a vertex set $\setcmd{V}$, an adjacency matrix $\matcmd{A}$, and a node feature matrix $\matcmd{X}$. Without loss of generality, let the vertices of $\mathcal{G}$ be consecutively labeled from 1 to $|\setcmd{V}|$, i.e., $\setcmd{V}=\{1,2,\dots,|\setcmd{V}|\}$.
We consider undirected and unweighted graphs. Thus, the adjacency matrix 
$\matcmd{A} \in \mathbb{R}^{\lvert \setcmd{V}\rvert \times \lvert \setcmd{V}\rvert}$ 
is binary: $\matcmd{A}_{ij}=1$ if there is an edge between nodes $i$ and $j$, and $\matcmd{A}_{ij}=0$ otherwise. Each node is assigned with a $d$-dimensional feature vector. The node feature matrix 
$\matcmd{X} \in \mathbb{R}^{\lvert \setcmd{V}\rvert \times d}$ 
arranges these feature vectors vertically such that the $i$-th row of $\matcmd{X}$, denoted as $\matcmd{X}^i \in \mathbb{R}^{d}$, is the feature vector of node $i$.

We focus on robust training of GNNs for semi-supervised node classification. It assumes that the nodes of a subset $\setcmd{V}_L \subset \setcmd{V}$ are assigned with labels in advance. Let $C> 1$ denote the number of node classes. Each node label $Y\in\{1,\dots,C\}$ can be alternatively represented as a one-hot vector $\veccmd{y}\in \{0,1\}^{C}$. Recall that node labels can be scarce and noisy in practice. In other words, (1) $|\setcmd{V}_L|$ is significantly smaller than $|\setcmd{V}|$, and (2) $\veccmd{y}^i$ might be incorrect for some of the nodes $i \in \setcmd{V}_L$. 
To tackle the node classification task, GNNs first compute a logit vector $\veccmd{o}^i_{\theta}\in \mathbb{R}^{C}$ for each node $i$, where $\theta$ is the model parameters. The probability of node $i$ belonging to class $j$ is then derived as:
\begin{equation}
    \veccmd{p}_{\theta}^{i,j}={\exp(\veccmd{o}^{i,j}_{\theta})} / {\sum_{k=1}^{C}\exp(\veccmd{o}^{i,k}_{\theta})},
\end{equation}
where $\veccmd{o}^{i,k}_{\theta}$ denotes the $k$-th entry of $\veccmd{o}^{i}_{\theta}$.

\eat{
For semi-supervised node classification, a subset $\setcmd{V}_L \subset \setcmd{V}$ of nodes are assigned with one-hot representation labels $y\in \{0,1\}^{C}$. Here $C$ denotes the number of the classes. Recall that we consider that node labels are scarce and noisy. In other words, (1) $|\setcmd{V}_L|$ is significantly less than $|\setcmd{V}|$, and (2) $y^i$ might be incorrect for some $i \in \setcmd{V}_L$. The goal of semi-supervised node classification is to map each unlabeled node $i \in \setcmd{V} \setminus \setcmd{V}_L$ to its true class label $y^i \in \{0,1\}^{C}$.
Let $\veccmd{o}^i_{\theta}\in \mathbb{R}^{C}$ denote the logit vector of node $i$ outputted a GCN parameterized by $\theta$. The probability of node $i$ belonging to class $j$ is derived as:
\begin{equation}
    \veccmd{p}_{\theta}^{i,j}={\exp(\veccmd{o}^{i,j}_{\theta})} / {\sum_{k=1}^{C}\exp(\veccmd{o}^{i,k}_{\theta})},
\end{equation}
where $\veccmd{o}^{i,k}_{\theta}$ is the $k$-th entry in $\veccmd{o}^{i}_{\theta}$.
}

\noindent\stitle{Two types of label noises}.
We consider two types of categorical label noises: {\it uniform} and {\it pair}. These two noise types are widely adopted in the literature~\cite{dai2021nrgnn,han2018co,malach2017decoupling}. Let $\epsilon$ denote the noise rate of node labels. For uniform noise, the labels have a probability of ${\epsilon}/{(C-1)}$ to be uniformly flipped to other classes. For pair noise, the labels probably make mistakes only  within very similar class~\cite{han2018co}, i.e., the labels have a probability of $\epsilon$ to be flipped to the paired class. \emph{In this work, we assume that the majority of labeled nodes have correct labels, i.e., the noise rate $\epsilon < 0.5$.}

%% file: files/4.methodology.tex
\section{Methodology}
\label{sec:methodology}

\begin{figure*}[t]
    \centering
    \includegraphics[width=0.9\textwidth]{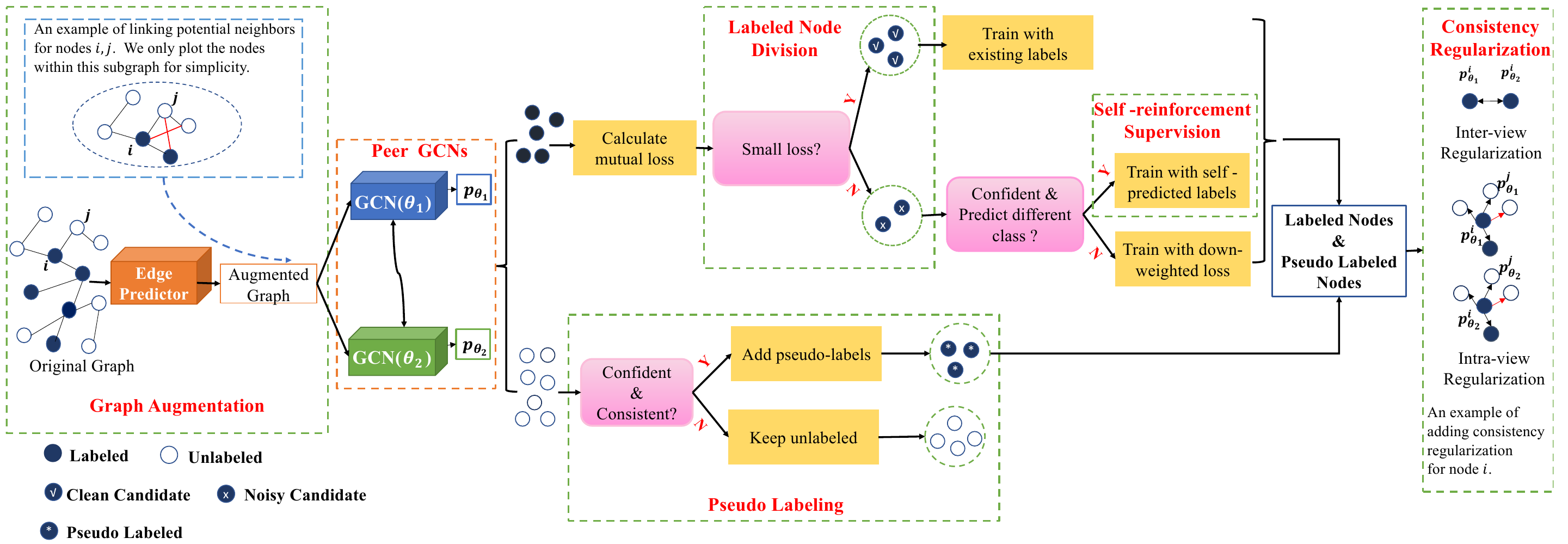}
    \vspace{-0.3cm}
    \caption{The overall architecture of \model. The red lines denote the added links. }
    \label{fig:overall}
    \vspace{-0.3cm}
\end{figure*}



\subsection{Overview}
\label{sec:overview}
Our new \model framework for \underline{R}obust \underline{T}raining of \underline{GNN}s with scarce and noisy node labels seeks to explicitly govern label noise to enable sufficient learning from clean labels while limiting the impact of noisy ones. An overview of \model is given in Fig.~\ref{fig:overall}. 

\model first augments the input graph by learning an edge predictor to infer potential links between labeled and unlabeled nodes (Sec.~\ref{sec:graphaugment}). These added links lead to more efficient message passing, thus mitigating the label scarcity issue.
The augmented graph is then fed to a pair of peer GCNs for explicit noise governance (Sec.~\ref{sec:noisegovern}). Based on the prediction of the two peer GCNs, nodes are classified into different categories, enabling nodes to choose supervision strategies adaptively. More specifically, labeled nodes are divided into clean candidate set  $\setcmd{V}_{cl}$ and noisy candidate set $\setcmd{V}_{ns}$ based on the small-loss criterion~\cite{han2018co} (Sec.~\ref{sec:LD}). The clean nodes are mainly supervised by their assigned labels. 
\model further identifies a subset $\setcmd{V}_{sr} \subseteq \setcmd{V}_{ns}$ of noisy label nodes whose predictions are confident but different from their assigned labels. It then reinforces the training of nodes in $\setcmd{V}_{sr}$ with their own predicted labels to alleviate noise propagation (Sec.~\ref{sec:SR}).

For the remaining nodes in $\setcmd{V}_{ns} \setminus \setcmd{V}_{sr}$, label-based training is performed with a down-weighted loss. 
Similar to the self-reinforce case, \model further generates pseudo-labels for those unlabeled nodes whose predictions are confident (Sec.~\ref{sec:PL}) in order to facilitate sufficient learning. Finally, it introduces the inter-view and intra-view regularizations to further prevent the model from overfitting to noise (Sec.~\ref{sec:CR}). We will summarize the different categories of nodes and their supervisions in Sec.~\ref{sec:totaltrain}.




\subsection{Graph Augmentation}
\label{sec:graphaugment}
Latent graph learning has shown to be beneficial in a number of graph learning tasks, owing to its flexibility to form graph structures optimized for specific tasks~\cite{kipf2016variational,pan2018adversarially,cosmo2020latent,zhao2021data,fatemi2021slaps,kazi2022differentiable}. Accordingly, we also utilize latent graph learning to densify the input graph as augmentation, to promote supervision propagation and alleviate label scarcity. Inspired by~\cite{dai2021nrgnn}, we choose to infer potential links between labeled and unlabeled nodes. The reasons are two-fold: 1) labeled nodes could provide direct data supervision to similar unlabeled ones, and 2) unlabeled nodes with pseudo-labels~(Sec.~\ref{sec:PL}) could facilitate model supervision on labeled ones as a supplement. 

\eat{The number of node pairs for potential links is $O(|\setcmd{V}|^2)$.}
To keep the extra computational cost of graph augmentation affordable, we decouple the process into candidate generation and link inference. For candidate generation, we retrieve the top-$K$ nearest labeled/unlabeled nodes for each unlabeled/labeled one and the distance is evaluated in the node feature space~\cite{fatemi2021slaps,dai2022towards}. The generation step can be conducted offline in advance and efficient approximation methods~\cite{halcrow2020grale,zhou2021informer} are available for large graphs.

Next, we train an encoder-decoder~\cite{kipf2016variational} edge predictor to infer links. A GCN~\cite{kipf2016semi} is utilized as the encoder to project nodes into a latent space based on both node feature and structure information:
\begin{equation}
\label{eq:graphaug}
    \matcmd{Z}=\mbox{GCN}(\matcmd{X},\matcmd{A}).
\end{equation}
We then adopt an inner-product decoder for link inference. More specifically, the \emph{predicted} edge weight $w_{ij}$ between nodes $i$ and $j$ is estimated as the non-negative cosine similarity between their node representations $\mathbf{z}^{i}$ and $\mathbf{z}^{j}$ learned by the encoder:
\begin{equation}
    w_{ij}=\max(\frac{\mathbf{z}^{i} \cdot \mathbf{z}^{j}}{\lVert \mathbf{z}^{i} \lVert \lVert \mathbf{z}^{j} \lVert  },0).
\end{equation}
We train the edge predictor with a negative-sampling based reconstruction objective:
\begin{equation}
\label{eq:rec}
    \setcmd{L}_{rec}=\sum_{i} \big(\sum_{j:\ \matcmd{A}_{ij}>0}(w_{ij}-\matcmd{A}_{ij})^{2}+ N_{neg} \cdot \mathbb{E}_{j^{'} \sim P_{n}}(w_{ij^{'}}-\matcmd{A}_{ij^{'}})^{2} \big),
\end{equation} 
where $N_{neg}$ is the number of negative samples for each node and $P_{n}$ is the distribution of negative samples (e.g., uniform). We employ negative sampling to improve computational efficiency and avoid bias towards negative node pairs. The adjacent matrix $\matcmd{\hat{A}}$ of the augmented graph is constructed as: 
\begin{equation}
\label{eq:augWeight}
\mathbf{\hat{A}}_{ij}=\left\{
\begin{aligned}
& 1 \hspace{5ex} & \mbox{if\ } \matcmd{A}_{ij}=1; \\
&w_{ij} \hspace{5ex} & \mbox{if\ } \matcmd{A}_{ij}=0,~j \in topK(i),~\mbox{and\ }w_{ij}>\tau; \\
& 0 & \mbox{otherwise}.
\end{aligned}
\right.
\end{equation}
Note that $topK(i)$ includes the retrieved nearest neighbor candidates for node $i$ and the threshold $\tau$ filters those unreliable ones.

Following~\cite{chen2019deep,dai2021nrgnn}, we train an end-to-end model by jointly learning node classification and graph reconstruction. 
The training process will be described in Sec.~\ref{sec:totaltrain}.


\subsection{Robust Training via Noise Governance}
\label{sec:noisegovern}

The core of \model 
lies in explicit label noise governance. To achieve this, \model feeds the augmented graph to a pair of peer GCNs that share the same architecture but have different parameters, $\theta_{1}$ and $\theta_{2}$. Based on the predictions of these two GCNs, says $\veccmd{p}_{\theta_{1}}$ and $\veccmd{p}_{\theta_{2}}$, \model classifies the nodes into different categories and further exploits different supervision strategies accordingly. As such, \model is able to effectively exploit the supervision of clean labels while limiting the impact of noisy ones.


\begin{figure}[]
\centering
\subfigtopskip=-3pt
\subfigcapskip=-3pt
\subfigure[5\% labeled nodes, uniform noise]{\includegraphics[width=0.46\linewidth]{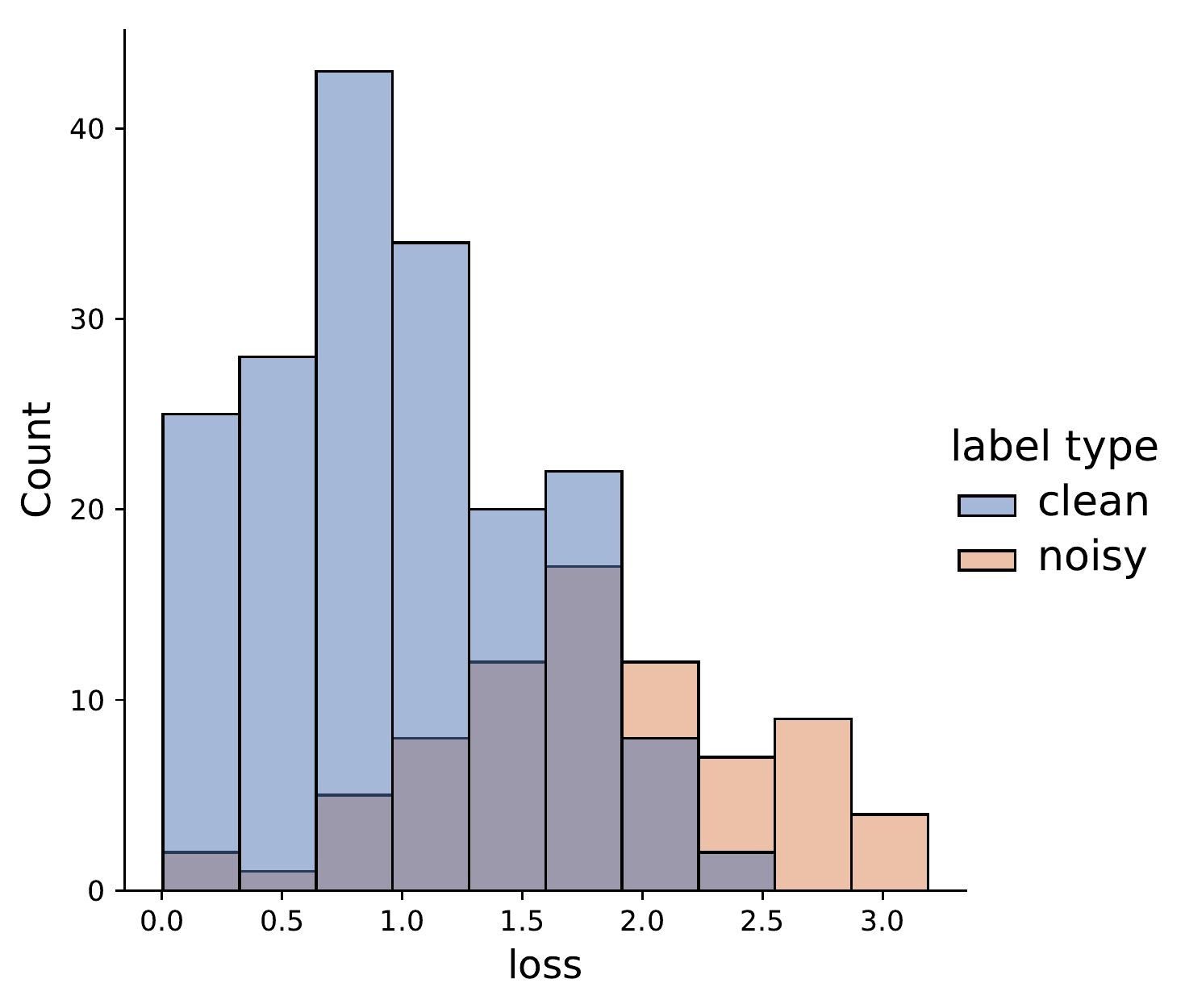}}
\subfigure[10\% labeled nodes, pair noise]{\includegraphics[width=0.46\linewidth]{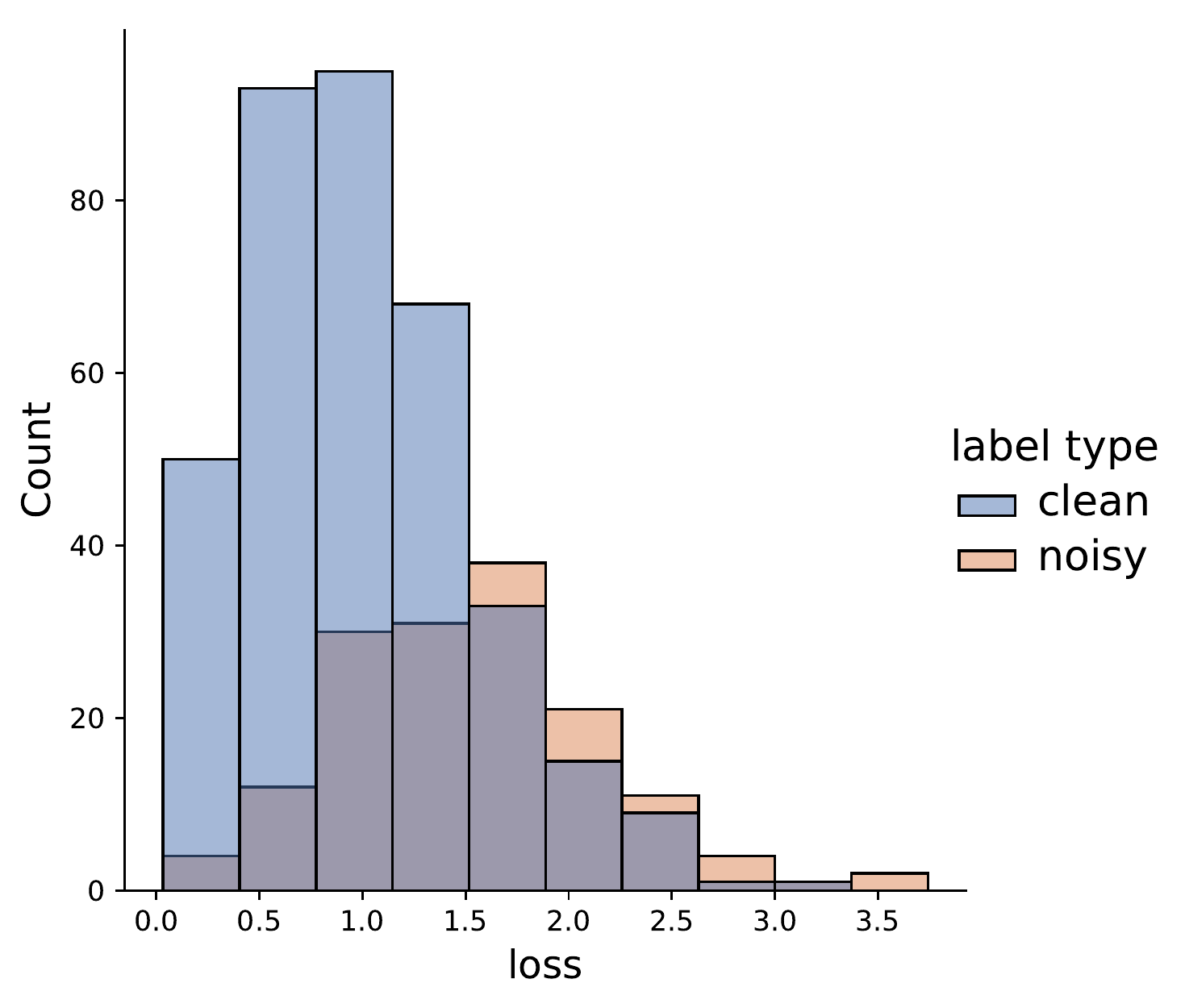}}
\vspace{-0.4cm}
\caption{
Loss distribution of the 40th epoch when training a GCN on BlogCatalog~\cite{wu2019net} with a noise rate $\epsilon=0.3$.}
\label{fig:lossdist}
\vspace{-0.4cm}
\end{figure}

\subsubsection{Labeled node division}
\label{sec:LD}
We first divide labeled nodes of $\setcmd{V}_{L}$ into clean and noisy candidate sets, $\setcmd{V}_{cl}$ and $\setcmd{V}_{ns}$.
Previous studies have observed that DNNs follow the pattern of learning clean labels first and then gradually overfitting to noisy labels~\cite{arpit2017closer}. This will lead to different loss distributions for clean and noisy nodes.
Fig.~\ref{fig:lossdist} also empirically verifies this phenomenon when training a GCN on graph data: The loss values of clean nodes are generally smaller than those of noisy nodes in the early stage. 
Inspired by the above observation, we adopt the small-loss criterion~\cite{han2018co} for labeled node division. Also, here we leverage the peer GCNs as different classifiers possess different abilities to exclude noises because they form diverse decision boundaries~\cite{han2018co,yu2019does}. We define the mutual loss derived from cross entropy as the loss metric for a node $i$, as:   
\begin{equation}
\label{eq:mutual}
\mathcal{L}_{mutual}^{i}= -\veccmd{y}^{i}\log({\veccmd{p}_{\theta_{1}}^{i}})-\veccmd{y}^{i}\log({\veccmd{p}_{\theta_{2}}^{i}})= -\veccmd{y}^{i}\log({\veccmd{p}_{\theta_{1}}^{i}}\cdot {\veccmd{p}_{\theta_{2}}^{i}}).
\end{equation}
The mutual loss measures prediction confidence, i.e., $\mathcal{L}_{mutual}^{i}$ has a low value if both the GCN classifiers correctly predict the class of node $i$. As clean labels are easier to learn, we expect that the mutual loss values for clean label nodes are smaller than noisy ones.  
 

Let $T_{max}$ denote the number of training epochs. In the $t$-th epoch, clean and noisy candidate sets are formally divided as follows: 
\begin{equation}
\label{eq:division}
\begin{aligned}
    th^{t}_{epoch}= & \ \mbox{Percentile}(\setcmd{L}^{i}_{mutual}, 100-\frac{50t}{T_{max}}),\\ 
    th^{t}_{avg} = & \ \mbox{Average}(\setcmd{L}^{i}_{mutual}), \\
    \setcmd{V}_{cl} = & \ \{i \mid \mathcal{L}^{i}_{mutual} < \max(th^{t}_{epoch}, th^{t}_{avg})\},\\
    \setcmd{V}_{ns} = & \ \setcmd{V}_{L} \setminus \setcmd{V}_{cl}.
\end{aligned}
\end{equation}

Notes: (1) $\mbox{Percentile}(\setcmd{L}, p)$ is the value below which $p\%$ of the loss values in $\setcmd{L}$ fall, (2) thresholds $th^{t}_{epoch}$ and $th^{t}_{avg}$ are the upper bounds of loss values for clean label nodes, (3) at least half of labeled nodes are classified as clean given a noise rate $\epsilon<0.5$, and (4) the threshold $th^{t}_{avg}$ further ensures that small-loss nodes are clean regardless of their relative ranks.

The training on nodes in $\setcmd{V}_{cl}$ is supervised by their assigned labels, with $\setcmd{L}^{i}_{mutual}$ being the classification loss. On the other hand, we leverage different supervisions for nodes in $\setcmd{V}_{ns}$, which will be discussed in the following subsections.
The above node division strategy has two advantages. 
First, it does not require hyper-parameters, thus avoiding elaborate parameter tuning. Second, it strikes a balance between sufficient training in the early stage and noise-resistant training in the later stage. 
At the early stage of training (e.g., the first few epochs), the model is underfitting and has not yet learned reliable patterns. By choosing a conservative (i.e., high) $th^{t}_{epoch}$, the majority of labeled nodes are classified as clean to guide model supervision. This can also be viewed as warm-up training. Besides, as mentioned before, the memorization effect of neural networks enables one to learn from clean labels first, despite the existence of noise. Thus, choosing a conservative $th^{t}_{epoch}$ would not hurt the model performance. 
When training goes on, we progressively exclude nodes with large loss from $\setcmd{V}_{cl}$, which prevents our model from overfitting to noise in the later stage. 


\subsubsection{Self-reinforcement supervision}
\label{sec:SR}
As the labels of nodes in $\setcmd{V}_{ns}$ are prone to be noisy, \model turns to other supervisions instead.
Note that properly trained deep models have the ability to predict the correct labels on their own. Thus, we give a chance for nodes in $\setcmd{V}_{ns}$ to reinforce the training with their own predicted labels. We identify a subset $\setcmd{V}_{sr} \subseteq \setcmd{V}_{ns}$ of nodes on which the predictions of the two peer GCNs are confident but different from their labels. Formally, $\setcmd{V}_{sr}$ includes the nodes that satisfy:
\begin{equation}
\label{def:disagreedconfident}
\begin{aligned}
   & {Z}^{i}=\arg\max_{c} \veccmd{p}_{\theta_{1}}^{i,c}=\arg\max_{c} \veccmd{p}_{\theta_{2}}^{i,c} \neq Y^{i}, \\
   & \sqrt{\veccmd{p}_{\theta_{1}}^{i,Z^{i}} \cdot \veccmd{p}_{\theta_{2}}^{i,Z^{i}}} > 1-\frac{(C-1)t}{C \cdot T_{max}},
\end{aligned}
\end{equation}
where $Y^{i}\in\{1,\dots,C\}$ denotes the label for node $i$. The threshold $1-\frac{(C-1)t}{C \cdot T_{max}}$ linearly decreases with the increment of $t$ and is finally reduced to $1/C$. In other words, we gradually allow more nodes to self-reinforce the training.

In each epoch, we update $\setcmd{V}_{sr}$ and further compute an adaptive training weight for each node $i \in \setcmd{V}_{sr}$:
\begin{equation}
\label{eq:selfcorrection}
\begin{aligned}
    \mu(i)=&  (\sqrt{\veccmd{p}_{\theta_{1}}^{i,Z^{i}} \cdot \veccmd{p}_{\theta_{2}}^{i,Z^{i}}})^{1-\frac{t}{T_{max}}},\\
    \setcmd{L}_{sr}^{i} =& -\mu(i)\veccmd{z}^{i}\log(\veccmd{p}^{i}_{\theta_{1}} \cdot \veccmd{p}^{i}_{\theta_{2}}),
\end{aligned}
\end{equation}
where $\veccmd{z}^{i} \in \{0,1\}^{C}$ is the corresponding one-hot representation of the predicted label $Z^{i}$, and $\mu(i)$ serves as an adaptive learning rate. 
Notes: (1) $\mu(i)$ relies on the node-wise prediction confidence. (2) The scaling exponent $(1-t/T_{max})$ gradually decreases to zero to enlarge $\mu(i)$ in the later training stage.


\subsubsection{Pseudo labeling}
\label{sec:PL}
Learning from pseudo-labels may further ease the training with scarce and noisy labels~\cite{dai2021nrgnn}. 
Similarly, we select a subset $\setcmd{V}_{pse} \subset \setcmd{V} \setminus \setcmd{V}_L$ of unlabeled nodes with confident and consistent predictions to generate pseudo-labels. Formally, $\setcmd{V}_{pse}$ includes the nodes that satisfy:
\begin{equation}
\label{eq:pse}
    \Tilde{Z}^{i}=\arg\max_{c} \veccmd{p}_{\theta_{1}}^{i,c}=\arg\max_{c} \veccmd{p}_{\theta_{2}}^{i,c},~~
    \sqrt{\mathbf{p}_{\theta_{1}}^{i,\Tilde{Z}^{i}}\cdot\veccmd{p}_{\theta_{2}}^{i,\Tilde{Z}^{i}}} > th_{pse},
\end{equation}
where $th_{pse}$ refers to the confidence threshold. The training on nodes in $\setcmd{V}_{pse}$ is then supervised by their pseudo-labels.

\subsubsection{Consistency regularization.}
\label{sec:CR}
For nodes in $\setcmd{V}_{L} \cup \setcmd{V}_{pse}$, we further add some regularization to prevent overfitting. Inspired by mutual learning~\cite{zhang2018deep}, we let the two peer GCNs mimic each other's prediction. In this way, they teach and learn from each other. Here we use $D_{KL}$~(Kullback-Liebler divergence) as the mimicry loss:
\begin{equation}
\label{eq:inter-view}
    \setcmd{L}_{inter}^{i} 
    = {D}_{KL}(\veccmd{p}_{\theta_{1}}^{i}\lVert \veccmd{p}_{\theta_{2}}^{i})+{D}_{KL}(\veccmd{p}_{\theta_{2}}^{i}\lVert \veccmd{p}_{\theta_{1}}^{i}),
\end{equation}

Moreover, we follow the local consistency (a.k.a. homophily~\cite{mcpherson2001birds}) assumption that linked nodes tend to belong to the same classes. This leads to a regularization term that enforces nodes to mimic neighbors' predictions within the same GCN:
\begin{equation}
\label{eq:neighbor}
    \setcmd{L}_{intra}^{i} 
    =\sum_{j} \frac{\veccmd{\hat{A}}_{ij}}{\sum_{k}{\veccmd{\hat{A}}}_{ik}} \big( 
        {D}_{KL}(\veccmd{p}^{j}_{\theta_{1}}\lVert \veccmd{p}^{i}_{\theta_{1}})
        + {D}_{KL}(\veccmd{p}^{j}_{\theta_{2}}\lVert \veccmd{p}^{i}_{\theta_{2}})
    \big).
\end{equation}

The final regularization loss is composed as:
\begin{equation}
\label{eq:reg}
    \setcmd{L}_{reg}^{i} = \setcmd{L}_{inter}^{i} + \setcmd{L}_{intra}^{i}.
\end{equation}

\subsection{The Overall Training Loss}
\label{sec:totaltrain}
Combining all together, the loss of labeled nodes can be unified as:
\begin{equation}
    \setcmd{L}_{labeled}=-\frac{1}{\lvert \setcmd{V}_{L} \rvert}\sum_{i\in V_{L}} 
   \big( \zeta(i)\veccmd{\hat{y}}^{i}\log(\veccmd{p}^i_{\theta_{1}} \cdot \veccmd{p}^i_{\theta_{2}}) + \lambda\setcmd{L}_{reg}^{i} \big),
\end{equation}
where
\begin{equation*}
\left\{
\begin{aligned}
& \zeta(i)=1,~\veccmd{\hat{y}}^{i}=\veccmd{y}^{i}  \hspace{10ex} & \mbox{if\ } i \in \setcmd{V}_{cl}; \\
&\zeta(i)=\mu(i),~\veccmd{\hat{y}}^{i}=\veccmd{z}^{i}  \hspace{10ex} & \mbox{if\ } i \in \setcmd{V}_{sr}; \\
&\zeta(i)=\gamma \in (0,1),~\veccmd{\hat{y}}^{i}=\veccmd{y}^{i}   & \mbox{if\ } i \in \setcmd{V}_{ns} \setminus \setcmd{V}_{sr}.
\end{aligned}
\right.
\end{equation*}
Note that $\gamma$ is a down-weighted factor for nodes with noisy labels and unconfident predictions. 

Similarly, the training loss for nodes with pseudo-labels is:
\begin{equation}
     \setcmd{L}_{pse}=-\frac{1}{\lvert \setcmd{V}_{pse} \rvert}\sum_{i\in\setcmd{V}_{pse}} \big( \veccmd{\Tilde{z}}^{i}\log(\veccmd{p}_{\theta_{1}}^{i} \cdot \veccmd{p}_{\theta_{2}}^{i})+\lambda\setcmd{L}_{reg}^{i} \big),
\end{equation}
where $\veccmd{\Tilde{z}^{i}} \in \{0,1\}^{C}$ is the corresponding one-hot representation of the predicted pseudo-label $\Tilde{Z}^{i}$~(Eq.~(\ref{eq:pse})).

Combining the graph reconstruction loss~(Eq.~(\ref{eq:rec})) given in Sec.~\ref{sec:graphaugment}, the total training loss is:
\begin{equation}
\label{ltotal}
    \setcmd{L}_{total}=\setcmd{L}_{labeled}+\setcmd{L}_{pse}+\alpha \cdot \setcmd{L}_{rec}.
\end{equation}

%% file: files/5.experiment.tex
\section{Experiments}
\label{sec:exp}
\begin{table}[]
\small
\vspace{-0.4cm}
\caption{Dataset statistics (the data is cited from \cite{jin2020graph,wu2019net}).} 
\label{table:stat}
\vspace{-0.3cm}
\begin{tabular}{c|cccc}
\hline
\textbf{Dataset}     & \textbf{Nodes} & \textbf{Edges} & \textbf{Features} & \textbf{Classes} \\ \hline
Cora        & 2,485           & 5,069           & 1,433              & 7                \\
Citeseer    & 2,110           & 3,668           & 3,703              & 6                \\
BlogCatalog & 5,196           & 171,743         & 8,189              & 6                \\ \hline
\end{tabular}%
\vspace{-0.4cm}
\end{table}

\begin{table*}[]
\caption{Comparison of node classification performance (test accuracy, mean$\pm$std, averaged by 5 runs) by different models. \eat{\%~ is reported with 5 different runs.} The best results are marked in bold, and the best baseline results are underlined.}
\vspace{-0.3cm}
\label{table:baseline}
\small
\begin{tabular}{c|c|cccc|cccc}
\hline
\multirow{2}{*}{\textbf{Dataset}} & \multirow{2}{*}{\textbf{Model}} & \multicolumn{4}{c|}{\textbf{Uniform Noise}}                                                                        & \multicolumn{4}{c}{\textbf{Pair Noise}}                                                             \\ \cline{3-10} 
                                  &                                 & \textbf{10\%}          & \textbf{20\%}                         & \textbf{30\%}          & \textbf{40\%}          & \textbf{10\%}          & \textbf{20\%}          & \textbf{30\%}          & \textbf{40\%}          \\ \hline
\multirow{8}{*}{Cora}             & GCN                             & 79.0$\pm$0.3          & 76.6$\pm$0.5                         & 68.8$\pm$0.9          & 60.1$\pm$3.0          & 79.8$\pm$0.4          & 75.0$\pm$1.4          & 67.0$\pm$1.0          & 59.3$\pm$0.6          \\
                                  & Co-teaching                     & 79.2$\pm$0.4          & 77.0$\pm$0.3                         & 69.5$\pm$0.7          & 62.7$\pm$1.8          & 80.6$\pm$0.5          & 76.7$\pm$0.8          & 67.7$\pm$0.7          & 63.8$\pm$0.8          \\
                                  & JoCoR                           & 79.3$\pm$0.2          & 76.9$\pm$0.3                         & 72.9$\pm$1.4          & 68.2$\pm$2.5          & 80.7$\pm$0.3          & 77.6$\pm$0.3          & 69.2$\pm$1.4          & 63.8$\pm$0.9          \\
                                  & SCE                              & 78.9$\pm$0.4          & 77.0$\pm$0.8                         & 69.8$\pm$0.9          & 60.0$\pm$3.4          & 80.5$\pm$0.2          & 77.4$\pm$1.0          & 67.8$\pm$2.4          & 57.8$\pm$1.3          \\
                                  & APL                             & 79.4$\pm$0.3          & 74.4$\pm$0.6                         & 69.2$\pm$1.4          & 59.5$\pm$3.2          & 80.0$\pm$0.5          & 75.2$\pm$2.3          & 67.4$\pm$0.9          & 59.8$\pm$2.2          \\
                                  & D-GNN                           & 79.6$\pm$0.1          & 74.7$\pm$0.6                         & 71.4$\pm$0.5          & 61.0$\pm$0.6          & 78.5$\pm$0.9          & 76.4$\pm$0.4          & 68.9$\pm$1.2          & 59.1$\pm$1.6          \\
                                  & NRGNN                           & {\ul 81.7$\pm$0.4}    & {\ul 80.4$\pm$1.2}                   & {\ul 76.7$\pm$0.7}    & {\ul 73.2$\pm$1.9}    & {\ul 81.7$\pm$0.4}    & {\ul 79.5$\pm$0.8}    & {\ul 75.4$\pm$1.1}    & {\ul 68.1$\pm$1.6}    \\
                                  & RTGNN (ours)                          & \textbf{82.6$\pm$0.7} & \textbf{80.8$\pm$0.2} & \textbf{79.9$\pm$1.0} & \textbf{78.1$\pm$1.1} & \textbf{82.9$\pm$0.3} & \textbf{80.9$\pm$0.6} & \textbf{77.4$\pm$0.6} & \textbf{72.1$\pm$1.4} \\ \hline
\multirow{8}{*}{Citeseer}         & GCN                             & 69.8$\pm$0.7          & 69.4$\pm$0.4                         & 63.5$\pm$1.2          & 55.0$\pm$1.0          & 69.2$\pm$0.5          & 67.0$\pm$0.8          & 58.5$\pm$2.1          & 57.5$\pm$1.6          \\
                                  & Co-teaching                     & 72.1$\pm$0.3          & 69.6$\pm$1.0                         & 69.0$\pm$1.6          & 56.9$\pm$1.4          & 72.2$\pm$0.5          & 69.3$\pm$1.3          & 60.1$\pm$1.7          & 57.6$\pm$2.4          \\
                                  & JoCoR                           & 72.2$\pm$0.3          & 71.3$\pm$0.7                         & {\ul 69.1$\pm$0.3}    & 58.1$\pm$2.3          & 72.1$\pm$0.6          & 69.6$\pm$0.4          & 64.4$\pm$2.5          & 58.1$\pm$1.6          \\
                                  & SCE                             & 70.2$\pm$0.4          & 69.4$\pm$0.7                         & 64.9$\pm$1.6          & 56.2$\pm$1.4          & 71.1$\pm$0.2          & 67.8$\pm$0.8          & 59.8$\pm$0.6          & 57.9$\pm$1.0          \\
                                  & APL                             & 70.5$\pm$0.6          & 69.6$\pm$0.6                         & 65.6$\pm$1.6          & 56.7$\pm$2.0          & 71.3$\pm$0.4          & 68.1$\pm$0.5          & 59.7$\pm$0.6          & 57.1$\pm$1.5          \\
                                  & D-GNN                           & 71.0$\pm$0.3          & 68.7$\pm$0.7                         & 63.1$\pm$0.5          & 54.5$\pm$0.9          & 70.0$\pm$0.5          & 69.5$\pm$0.5          & 62.8$\pm$0.6          & 56.4$\pm$4.0          \\
                                  & NRGNN                           & {\ul 72.6$\pm$0.8}    & {\ul 72.4$\pm$0.8}                   & 68.9$\pm$1.1    & {\ul 63.5$\pm$1.3}    & {\ul 72.8$\pm$0.5}    & {\ul 70.4$\pm$0.8}    & {\ul 65.0$\pm$1.3}    & {\ul 58.4$\pm$3.4}    \\
                                  & RTGNN (ours)                           & \textbf{74.5$\pm$0.6} & \textbf{74.1$\pm$0.6}                & \textbf{70.9$\pm$0.7} & \textbf{66.0$\pm$1.7} & \textbf{74.6$\pm$0.5} & \textbf{71.8$\pm$1.0} & \textbf{67.8$\pm$0.8} & \textbf{62.1$\pm$1.6} \\ \hline
\multirow{8}{*}{BlogCatalog}      & GCN                             & 69.9$\pm$0.7          & 66.4$\pm$0.8                         & 65.9$\pm$0.9          & 65.3$\pm$0.9          & 69.9$\pm$0.4          & 62.9$\pm$0.5          & 58.3$\pm$0.8          & 57.3$\pm$1.2          \\
                                  & Co-teaching                     & 70.9$\pm$0.3          & 69.6$\pm$0.5                         & 68.7$\pm$0.6          & 66.3$\pm$0.6          & 70.2$\pm$0.7          & 65.3$\pm$0.9          & 58.8$\pm$1.0          & 57.8$\pm$1.5          \\
                                  & JoCoR                           & 70.9$\pm$0.4          & 69.6$\pm$0.4                         & 69.3$\pm$1.1          & 66.3$\pm$1.1           & {\ul 70.5$\pm$0.7}    & 66.3$\pm$1.1          & 59.5$\pm$1.1          & 58.3$\pm$2.2          \\
                                  & SCE                             & 70.7$\pm$0.5          & 68.7$\pm$0.8                         & 67.5$\pm$0.9          & 66.5$\pm$1.0          & 69.4$\pm$0.7          & 64.3$\pm$1.4          & 60.6$\pm$1.0          & 57.0$\pm$0.9          \\
                                  & APL                             & 70.9$\pm$0.5          & 69.2$\pm$0.4                         & 68.7$\pm$0.6          & 65.8$\pm$0.9          & 70.3$\pm$0.6          & {\ul 68.4$\pm$0.8}    & {\ul 61.8$\pm$1.1}    & 58.0$\pm$2.7          \\
                                  & D-GNN                           & 70.7$\pm$0.3          & 67.9$\pm$0.6                         & 67.5$\pm$0.5          & 65.8$\pm$0.7          & 70.1$\pm$0.5          & 67.1$\pm$1.1          & 61.5$\pm$1.3          & {\ul 58.4$\pm$1.6}    \\
                                  & NRGNN                           & {\ul 71.1$\pm$0.4}    & {\ul 70.3$\pm$1.0}                   & {\ul 69.5$\pm$0.9}    & {\ul 67.0$\pm$1.3}    & 69.9$\pm$0.7          & 67.9$\pm$1.0          & 60.7$\pm$1.0          & 58.3$\pm$2.2          \\
                                  & RTGNN (ours)                           & \textbf{71.2$\pm$0.3} & \textbf{70.9$\pm$0.4}                & \textbf{70.7$\pm$0.4} & \textbf{68.0$\pm$1.1} & \textbf{71.1$\pm$0.4} & \textbf{70.3$\pm$0.8} & \textbf{65.2$\pm$1.0} & \textbf{63.6$\pm$1.5} \\ \hline
\end{tabular}%
\vspace{-0.3cm}
\end{table*}

\subsection{Experiment Setup}
\textbf{Datasets.}
We conduct experiments on two  citation  datasets~\cite{jin2020graph} (Cora and Citeseer) and a social network dataset BlogCatalog~\cite{wu2019net} to evaluate the performance of \model. The dataset statistics are summarized in Table~\ref{table:stat}. 
To study the robustness of GNNs with scarce and noisy labels, each dataset is randomly split into 5\%, 15\%, and 80\% for training, validation, and test, respectively. Moreover, following~\cite{dai2021nrgnn}, we further randomly corrupt a fraction of labels, says $\epsilon$, in the training and validation sets. Recall in Sec.~\ref{sec:preliminaries} that we consider two types of noises. Each dataset thus has two corrupted versions corresponding to uniform and pair noises, respectively.

\noindent\textbf{Implementation details.}
We adopt a 2-layer GCN~\cite{kipf2016semi} with 128 hidden units as the backbone GNN model. 
For graph augmentation, we use a GCN with 64 hidden units  as the encoder of edge predictor, fix $N_{neg}=100$ for negative sampling, and set threshold $\tau=0.05$ on all datasets. The regularization loss weight $\lambda=0.1$ for all datasets. 

%
Moreover, we search  the  $K$ for nearest neighbor candidate~(the $topK$ operation in Eq.~(\ref{eq:augWeight})) in \{25, 50, 100\}, the pseudo label threshold $th_{pse}$ in \{0.7, 0.8, 0.9, 0.95\}, the reconstruction loss weight $\alpha$ in \{0.03, 0.1, 0.3, 1\}, and the down-weighted factor $\gamma$ of noisy samples in \{0.01, 0.1\}  based on the validation performance. 
Finally, we train our model for a total of 200 epochs with a learning rate of 0.001 and a weight decay of 5e-4.  
To avoid over-fitting, we also apply dropout with a dropout rate of 0.5. 
In the cases when quantitative measurements are reported, the test was repeated over 5 times and the average is reported\footnote{ \textcolor{red}{Codes are available at https://github.com/GhostQ99/RobustTrainingGNN.}}. \textbf{Notice that both GCNs could be used for inference, and in the experiments, we use the first one.}

\subsection{Performance Comparison with Baselines}
We first compare our \model with the following state-of-the-art robust learning baselines to evaluate the overall performance.
\begin{itemize}[leftmargin=*]
\item \textbf{GCN~\cite{kipf2016semi}} is a popular GNN model based on first-order approximate spectral convolution. 
\item \textbf{Co-teaching~\cite{han2018co}} trains paired peer networks such that each network excludes a proportion (based on the given noise rate) of large-loss nodes and uses the rest  to update the parameters of the peer network. 

\item \textbf{JoCoR~\cite{wei2020combating}} improves Co-teaching by jointly training and discarding a proportion of large-loss nodes. \eat{that are considered noisy.} It also adopts the paired network architecture and assumes a given noise rate. 

\item \textbf{SCE~\cite{wang2019symmetric}} enhances noise robustness by adding a noise-tolerant reverse cross entropy term to the cross entropy loss. 

\item \textbf{APL~\cite{ma2020normalized}} finds that normalization helps improve noise robustness of loss functions. It then categorizes robust loss functions into active and passive and combines them for training. In particular, it leverages a combination of normalized cross entropy and reverse cross entropy. 

\item \textbf{D-GNN~\cite{nt2019learning}} introduces backward loss correction~\cite{patrini2017making} to GNNs, which multiplies the cross entropy loss with the inverse of the noise matrix to alleviate loss distribution bias.

\item \textbf{NRGNN~\cite{dai2021nrgnn}} adopts two asymmetric GCNs for edge prediction and pseudo-label mining. It emphasizes on extracting and broadcasting more potentially correct supervision on graphs while neglecting proactive governance of noises. 

\end{itemize}
We use the official implementations of these baselines.
For a fair comparison, all the methods adopt a 2-layer GCN with 128 hidden units as the backbone model.
We evaluate the node classification accuracy of all methods on three datasets with two types of noises and varying noise rates. The results are reported in Table~\ref{table:baseline} and we observe the following from these results.
\begin{itemize} [leftmargin=*]
\item Compared with GCN~\cite{kipf2016semi}, loss-centric methods SCE, APL, and D-GNN perform even worse in some scenarios (e.g., SCE on Cora with 40\% pair noise, APL on Cora with 40\% uniform noise, and D-GNN on Citeseer with 40\% pair noise). It indicates that leveraging a single robust loss function or loss correction method is insufficient. Differently, \model explicitly divides nodes into several categories and adopts different training strategies accordingly.


\item Compared with GCN, both Co-teaching and JoCoR yield stable and superior performance. This verifies that co-training two networks can mutually suppress the effects of noise, inspiring us to borrow the peer network architecture in our model. It also validates that performing explicit governance of clean and noisy labels is beneficial.
\item JoCoR attains an overall better performance than Co-teaching by jointly training two networks together. This motivates us to fuse the training of two peer networks. Moreover, our \model works even better, because we further leverage the model's own ability  to rectify the noisy labels. Besides, we complement the training with confident and consistent pseudo labels.

\item NRGNN links unlabeled nodes with similar labeled nodes and adds pseudo-labels to mitigate the negative effect of noise. This strategy helps to bring effective supervision to unlabeled nodes. However, it might not perform well under higher noise rates since a large number of erroneous supervisions are propagated to a wide scope. 
To address this issue, we propose explicit noise governance to identify potentially noisy labels and alleviate their impacts. We consider our labeled node division as a primary module in \model, with other types of supervisions being introduced to further boost the performance. \textbf{Overall, \model consistently performs better than all baselines on three datasets with both types of noises and varying noise rates.}
\end{itemize}

\subsection{Impacts of Training Label Rates}

\begin{figure}[]
\centering
\subfigtopskip=-3pt
\subfigcapskip=-3pt
\subfigure[30\% Uniform Noise ]{\includegraphics[width=0.47\linewidth]{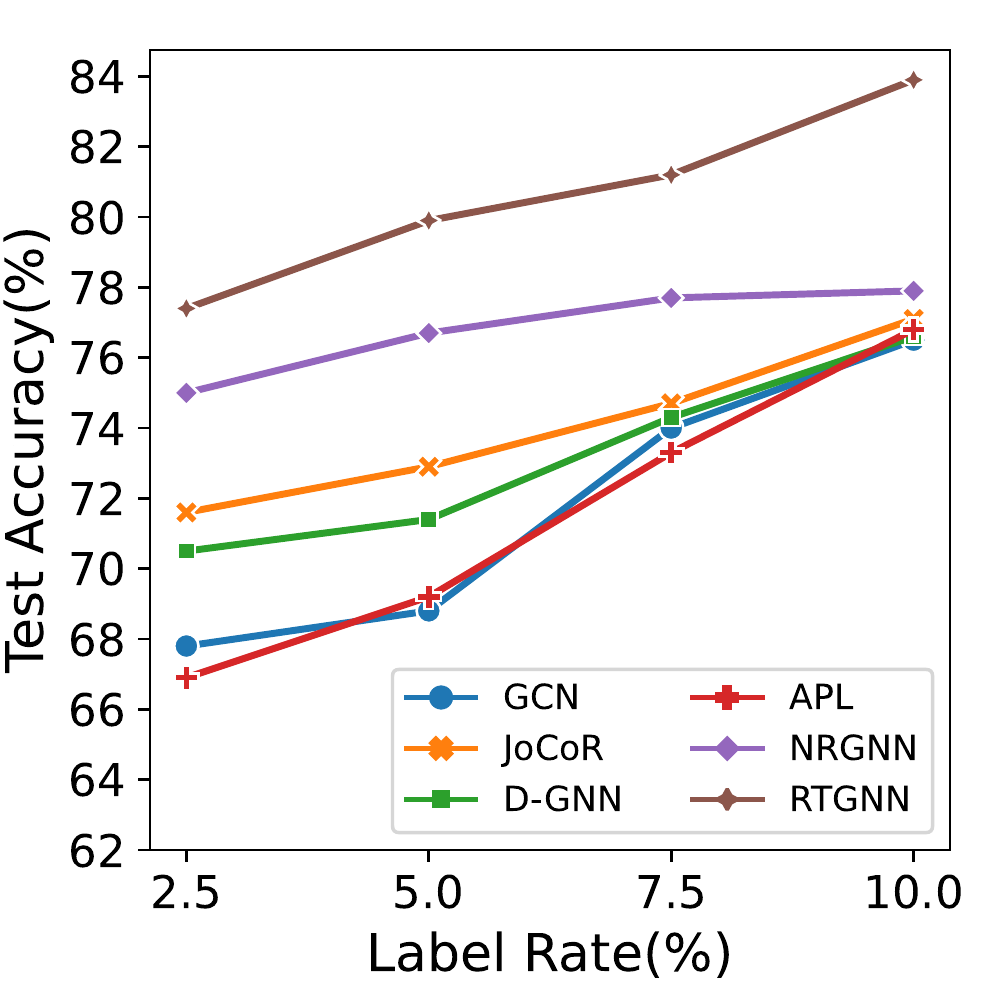}
\label{fig:labelCora:a}}
\subfigure[30\% Pair Noise]{\includegraphics[width=0.47\linewidth]{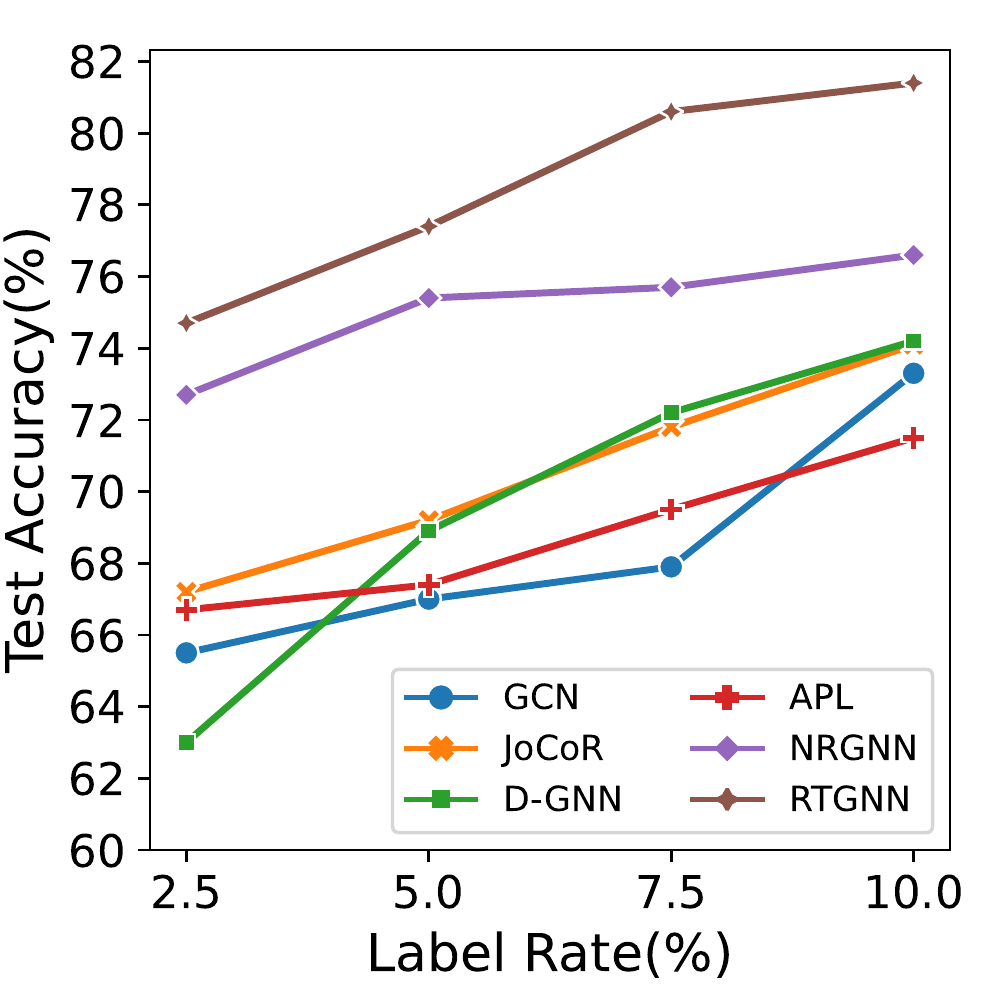}
\label{fig:labelCora:b}}
\vspace{-0.4cm}
\caption{Impacts of training label rates on Cora. \eat{with two types of noises and varying noise rates}}
\label{fig:labelCora}
\vspace{-0.4cm}
\end{figure}

\begin{figure}[]
\centering
\subfigtopskip=-3pt
\subfigcapskip=-3pt
\subfigure[30\% Uniform Noise]{\includegraphics[width=0.47\linewidth]{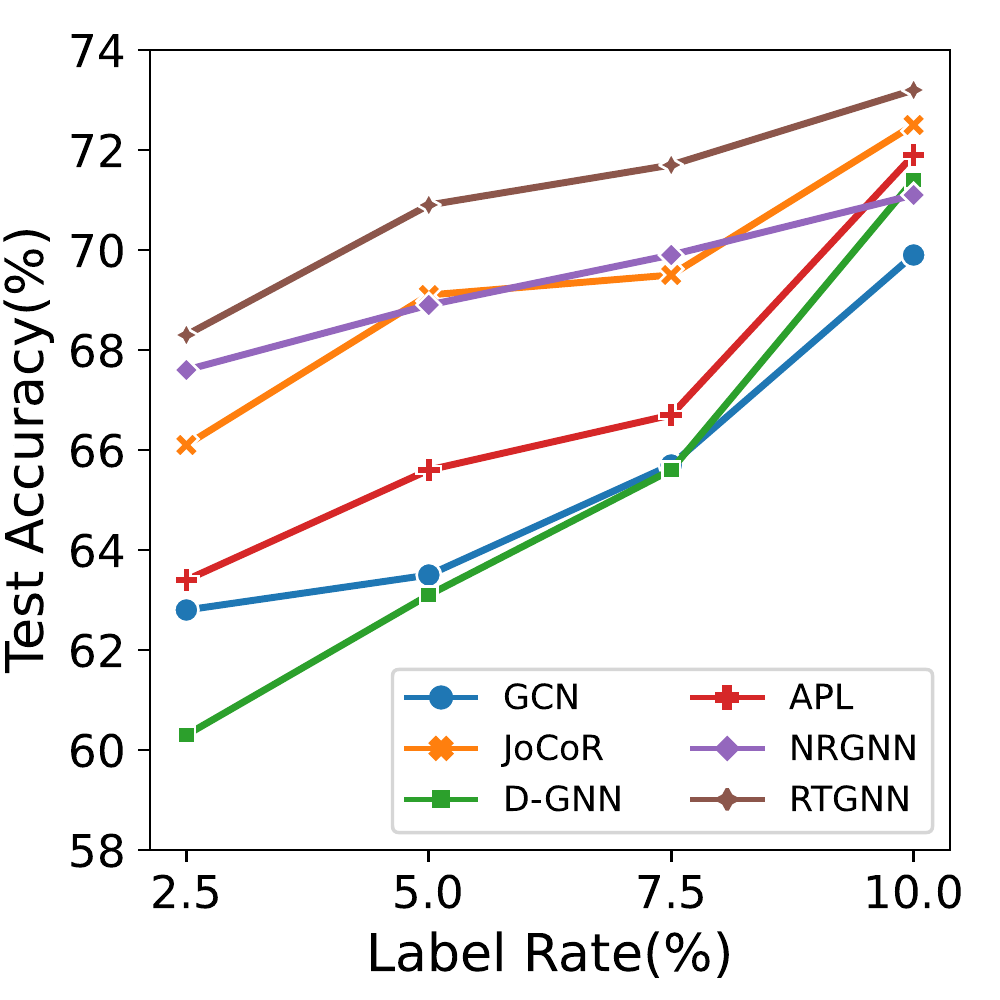}
\label{fig:labelcite:a}}
\subfigure[30\% Pair Noise]{\includegraphics[width=0.47\linewidth]{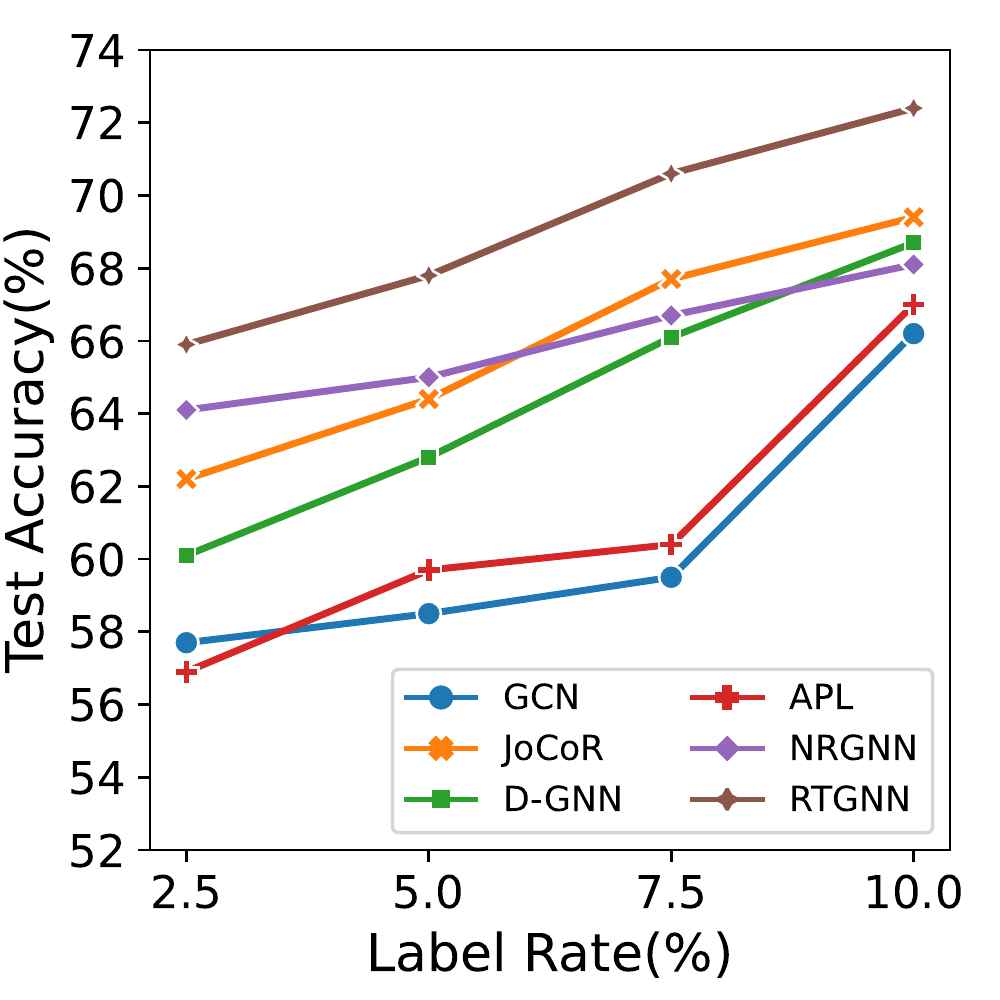}
\label{fig:labelcite:b}}
\vspace{-0.4cm}
\caption{Impacts of training label rates on Citeseer.}
\label{fig:labelciteseer}
\vspace{-0.4cm}
\end{figure}
We next examine how our \model performs with different fractions of training labels. For this purpose, we split the datasets into $x\%$, $(20-x)\%$, and 80\% for training, validation, and test, respectively.
Since we focus on training GNNs with sparse and noisy labels, we vary $x$ from 2.5 to 10 
and evaluate the test node classification accuracy. The results on Cora and Citeseer with 30\% noise rate are reported in Fig.~\ref{fig:labelCora} and Fig.~\ref{fig:labelciteseer}. Note that results are similar with other noise rates and we observe the following.
\begin{itemize}[leftmargin=*]
    \item Our proposed \textbf{\model} consistently outperforms all baselines. Even under an extremely low label rate ($x$ = 2.5\%), our model achieves favorable performances by leveraging the supervision of clean labels while restricting that of noisy ones. 
    \item Compared using the low (2.5\%) and high (10\%) label rates, we observe that our model makes a bigger improvement upon the competitive NRGNN baseline with high (10\%) label rates. As we mentioned before, without explicit noise governance, NRGNN mix-ups the propagation of clean and noisy labels in graphs, thus being severely affected by the increasing size of noisy labels.
\end{itemize}

\subsection{Ablation Study}
\label{sec:ablation}
\subsubsection{Ablation study on graph augmentation}
\begin{table}[]
\small
\caption{Ablation study of graph augmentation. ``w/ GA'' means plugging the graph augmentation module.}
\vspace{-0.3cm}
\label{table:abga}
\resizebox{0.975\linewidth}{!}{%
\begin{tabular}{c|c|cc|cc}
\hline
\multirow{2}{*}{\textbf{Dataset}}                                        & \multirow{2}{*}{\textbf{Variants}} & \multicolumn{2}{c|}{\textbf{Uniform Noise}}         & \multicolumn{2}{c}{\textbf{Pair Noise}}             \\ \cline{3-6} 
                                                                         &                                    & \textbf{30\%}         & \textbf{40\%}         & \textbf{30\%}         & \textbf{40\%}         \\ \hline
\multirow{6}{*}{\begin{tabular}[c]{@{}c@{}}Cite-\\ seer\end{tabular}}    & CE                                 & 63.5$\pm$1.2          & 55.0$\pm$1.0          & 58.5$\pm$2.1          & 57.5$\pm$1.6          \\
                                                                         & CE w/ GA                           & \textbf{67.9$\pm$1.6} & \textbf{62.8$\pm$1.9} & \textbf{64.6$\pm$3.2} & \textbf{58.6$\pm$3.2} \\ \cline{2-6} 
                                                                         & SCE                                & 64.9$\pm$1.6          & 56.2$\pm$1.4          & 59.8$\pm$0.6          & 57.9$\pm$1.0          \\
                                                                         & SCE w/ GA                          & \textbf{69.2$\pm$1.7} & \textbf{63.9$\pm$1.0} & \textbf{64.8$\pm$1.4} & \textbf{60.4$\pm$3.0} \\ \cline{2-6} 
                                                                         & APL                                & 65.6$\pm$1.6          & 56.7$\pm$2.0          & 59.7$\pm$0.6          & 57.1$\pm$1.5          \\
                                                                         & APL w/ GA                          & \textbf{69.5$\pm$1.2} & \textbf{63.2$\pm$0.9} & \textbf{64.3$\pm$1.2} & \textbf{59.5$\pm$1.6} \\ \hline
\multirow{6}{*}{\begin{tabular}[c]{@{}c@{}}Blog-\\ Catalog\end{tabular}} & CE                                 & 65.9$\pm$0.9          & 65.3$\pm$0.9          & 58.3$\pm$0.8          & 57.3$\pm$1.2          \\
                                                                         & CE w/ GA                           & \textbf{68.0$\pm$0.4} & \textbf{65.9$\pm$1.9} & \textbf{59.8$\pm$0.7} & \textbf{58.1$\pm$1.7} \\ \cline{2-6} 
                                                                         & SCE                                & 67.5$\pm$0.9          & 66.5$\pm$1.0          & 60.6$\pm$1.0          & 57.0$\pm$0.9          \\
                                                                         & SCE w/ GA                          & \textbf{69.8$\pm$1.0} & \textbf{67.1$\pm$0.3} & \textbf{62.6$\pm$0.7} & \textbf{58.7$\pm$0.5} \\ \cline{2-6} 
                                                                         & APL                                & 68.7$\pm$0.6          & 65.8$\pm$0.9          & 61.8$\pm$1.1          & 58.0$\pm$2.7          \\
                                                                         & APL w/ GA                          & \textbf{69.6$\pm$1.0} & \textbf{67.3$\pm$0.6} & \textbf{62.6$\pm$1.0} & \textbf{59.1$\pm$0.5} \\ \hline
\end{tabular}%
}
\vspace{-0.3cm}
\end{table}
To validate the effect of the graph augmentation module, we plug it into a GCN. In addition to the cross entropy~(CE) loss, various loss functions are also tested. 
As in Eq.~(\ref{ltotal}), we train the augmented baselines by jointly learning node classification and graph reconstruction. Similarly, we fine-tune the loss weight $\alpha$ and $K$ for the nearest neighbor candidate. We set the label rate as $x$ = 5\%, and report the results in Table~\ref{table:abga}.  We observe the following.
\begin{itemize}[leftmargin=*]
\item The graph augmentation module  boosts the training of GCN with scarce and noisy labels using various loss functions.
\item 
A sparse graph (Citeseer) gains more benefits than a denser one (BlogCatalog). On a dense graph, unlabeled nodes receive sufficient but probably wrong supervision from labeled neighbors. Without an appropriate strategy to govern label noise, the chance for unlabeled nodes to receive correct supervision isn't significantly increased, thus yielding limited performance improvement. This suggests that developing an effective noise governance strategy can largely help deal with graphs of different densities.
\end{itemize}

\subsubsection{Ablation study on noise governance}
We further investigate how each sub-module in our noise governance strategy contributes to the overall performance of \model. For simplicity, we denote these sub-modules as follows.
\begin{itemize}[leftmargin=*]
    \item \textbf{LD:} Labeled node division (introduced in Sec.~\ref{sec:LD}).
     \item \textbf{SR:} Self-reinforcement supervision (introduced in Sec.~\ref{sec:SR}). It is an operation following labeled node division.
    \item \textbf{PL:} Pseudo labeling (introduced in Sec.~\ref{sec:PL}).
        \item \textbf{CR:} Consistency regularization (introduced in Sec~\ref{sec:CR}).
\end{itemize}
\begin{table}[]
\small
\caption{Ablation study of noise governance.}
\vspace{-0.3cm}
\label{table:abng}
\resizebox{0.975\linewidth}{!}{%
\begin{tabular}{c|c|cc|cc}
\hline
\multirow{2}{*}{\textbf{Dataset}}                                        & \multirow{2}{*}{\textbf{Variants}} & \multicolumn{2}{c|}{\textbf{Uniform Noise}}         & \multicolumn{2}{c}{\textbf{Pair Noise}}             \\ \cline{3-6} 
                                                                         &                                    & \textbf{30\%}         & \textbf{40\%}         & \textbf{30\%}         & \textbf{40\%}         \\ \hline
\multirow{5}{*}{\begin{tabular}[c]{@{}c@{}}Cite-\\ seer\end{tabular}}    & w/o LD, SR                         & 69.5$\pm$0.8          & 64.0$\pm$1.3          & 66.2$\pm$0.7          & 59.7$\pm$1.3          \\
                                                                         & w/o SR                             & 70.3$\pm$0.9          & 65.1$\pm$1.3          & 67.2$\pm$0.7          & 61.1$\pm$1.5          \\
                                                                         & w/o PL                             & 70.5$\pm$1.2          & 65.7$\pm$1.5          & 67.3$\pm$1.7          & 61.4$\pm$1.8          \\
                                                                         & w/o CR                             & 70.6$\pm$1.0          & 65.4$\pm$2.1          & 67.3$\pm$1.2          & 61.2$\pm$1.9          \\
                                                                         & All                                & \textbf{70.9$\pm$0.7} & \textbf{66.0$\pm$1.7} & \textbf{67.8$\pm$0.8} & \textbf{62.1$\pm$1.6} \\ \hline
\multirow{5}{*}{\begin{tabular}[c]{@{}c@{}}Blog-\\ Catalog\end{tabular}} & w/o LD, SR                         & 69.2$\pm$0.5          & 67.1$\pm$0.6          & 61.4$\pm$0.5          & 58.5$\pm$1.2          \\
                                                                         & w/o SR                             & 70.1$\pm$0.4          & 67.5$\pm$0.9          & 64.5$\pm$1.2          & 62.0$\pm$1.3          \\
                                                                         & w/o PL                             & 70.4$\pm$0.4          & 67.5$\pm$1.3          & 64.6$\pm$1.1          & 62.9$\pm$2.0          \\
                                                                         & w/o CR                             & 70.3$\pm$0.2          & 67.8$\pm$1.1          & 64.4$\pm$1.4          & 62.5$\pm$1.7          \\
                                                                         & All                                & \textbf{70.7$\pm$0.4} & \textbf{68.0$\pm$1.1} & \textbf{65.2$\pm$1.0} & \textbf{63.6$\pm$1.5} \\ \hline
\end{tabular}%
}
\vspace{-0.5cm}
\end{table}

We set the label rate as $x$ = 5\%, and present the results in Table~\ref{table:abng}. From these results, we observe the followings. 
\begin{itemize}[leftmargin=*]
    \item The \model model without our labeled node division (LD) performs the worst. In particular, under 40\% pair noise, it drops by 2.4\% on Citeseer and by 5.1\% on BlogCatalog. This indicates the importance of dividing labeled nodes into clean and noisy candidate sets and leveraging different training strategies for them. Meanwhile, the effectiveness of our designed labeled node division strategy is also validated.
    \item Self-reinforcement supervision (SR) further exploits the neural network's ability for reinforcing itself with rectified labels, thus attaining more accurate supervision.
    \item Pseudo labeling (PL) enhances the performance because it serves as a kind of complementary supervision.
   \item Consistency regularization (CR) further prevents nodes from overfitting to noise by ensembling two classifiers to exclude noise together and exploiting local homogeneity in the graph.
\end{itemize}

\subsection{Sensitivity Analysis of Hyper-parameters}

\begin{figure}[]
\centering
\subfigtopskip=-3pt
\subfigcapskip=-3pt
\subfigure[Impact of $\alpha$ on Citeseer]{\includegraphics[width=0.47\linewidth]{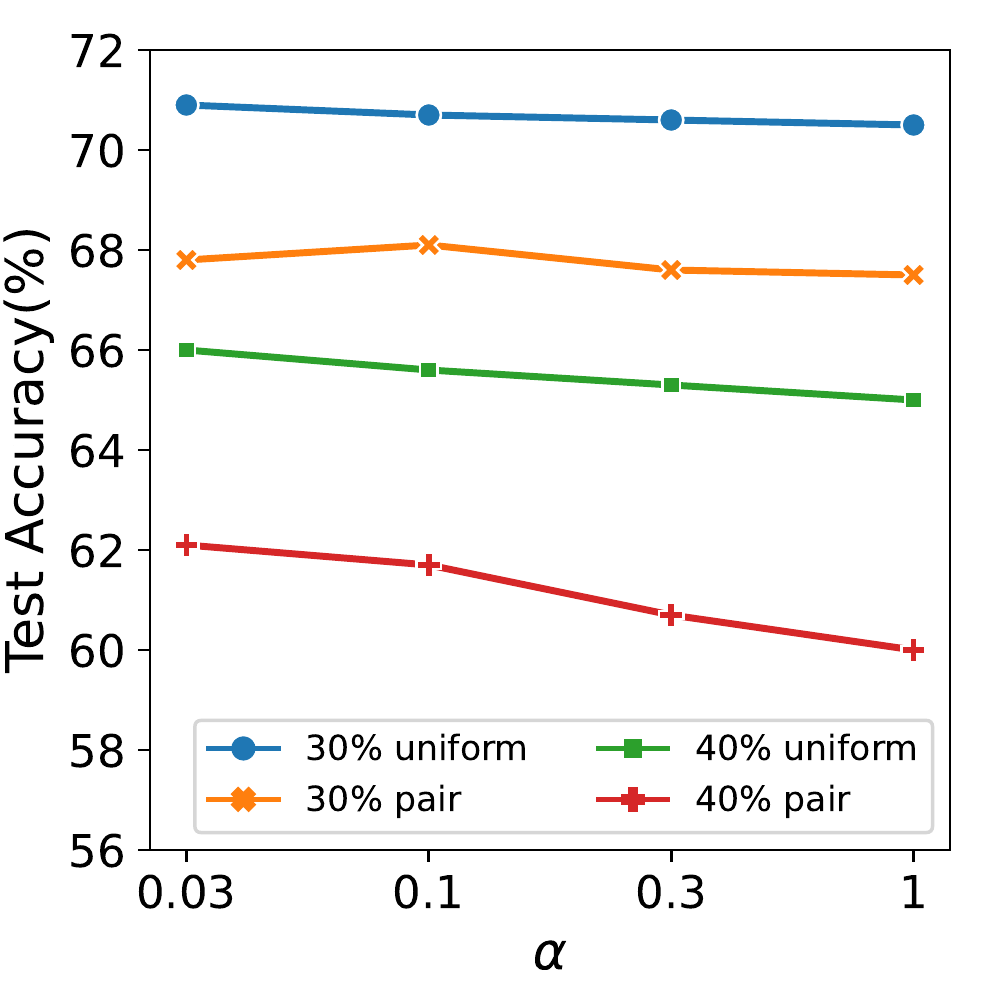}
\label{fig:hyperalpha:a}}
\subfigure[Impact of $\alpha$ on BlogCatalog]{\includegraphics[width=0.47\linewidth]{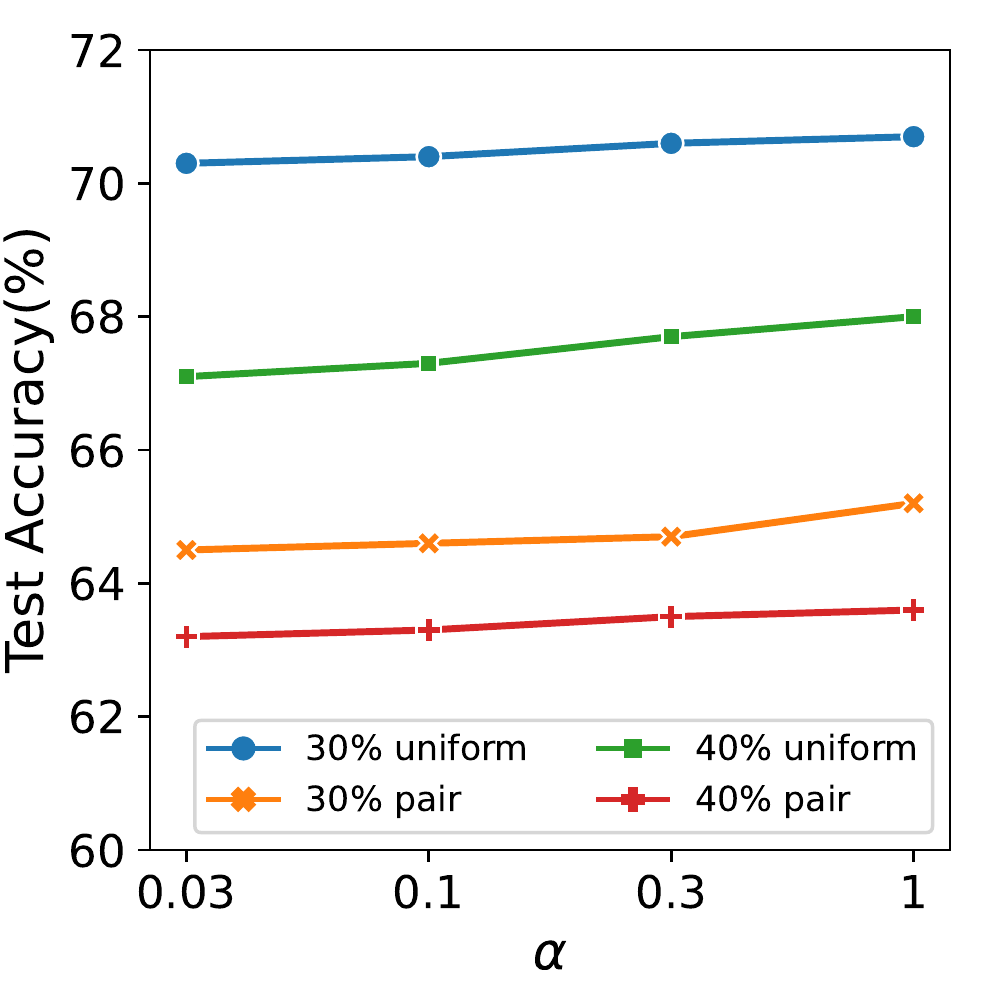}
\label{fig:hyperalpha:b}}
\subfigure[Impact of $\tau$ on Citeseer]{\includegraphics[width=0.47\linewidth]{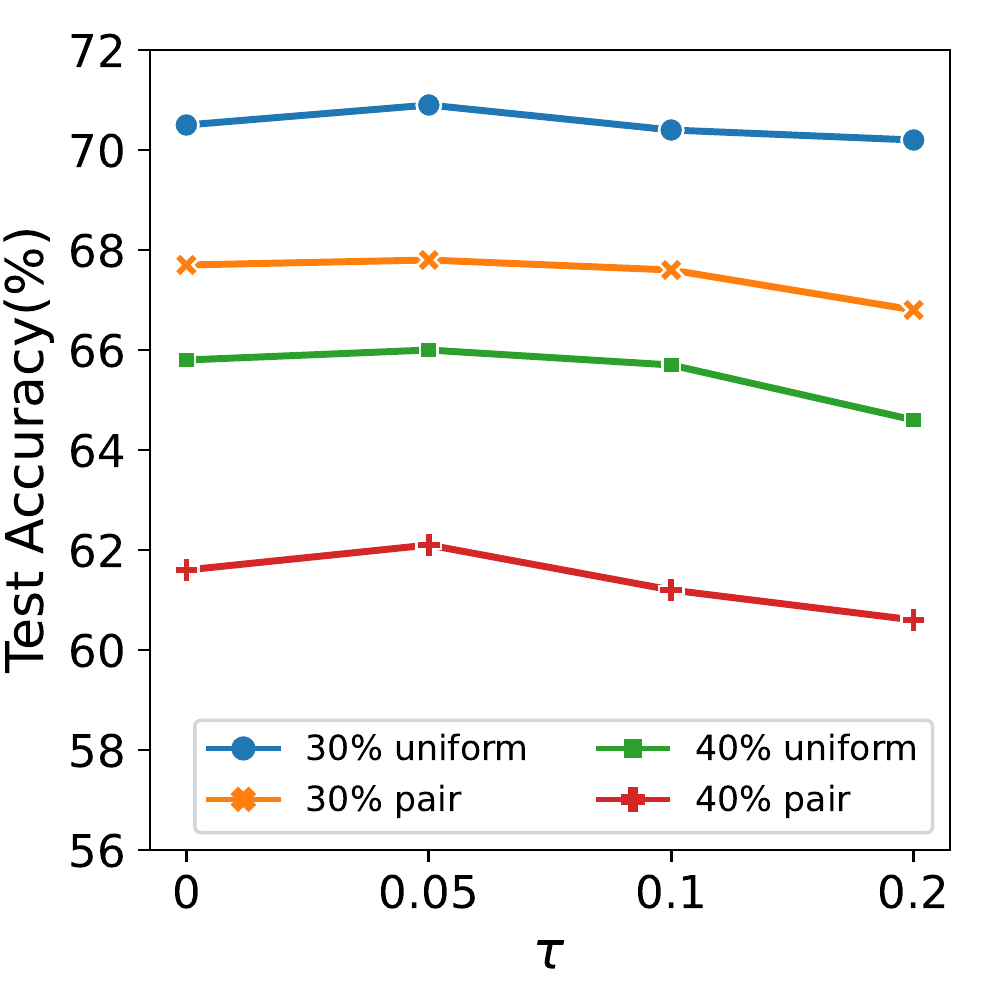}
\label{fig:hypertsmall:a}}
\subfigure[Impact of $\tau$ on BlogCatalog]{\includegraphics[width=0.47\linewidth]{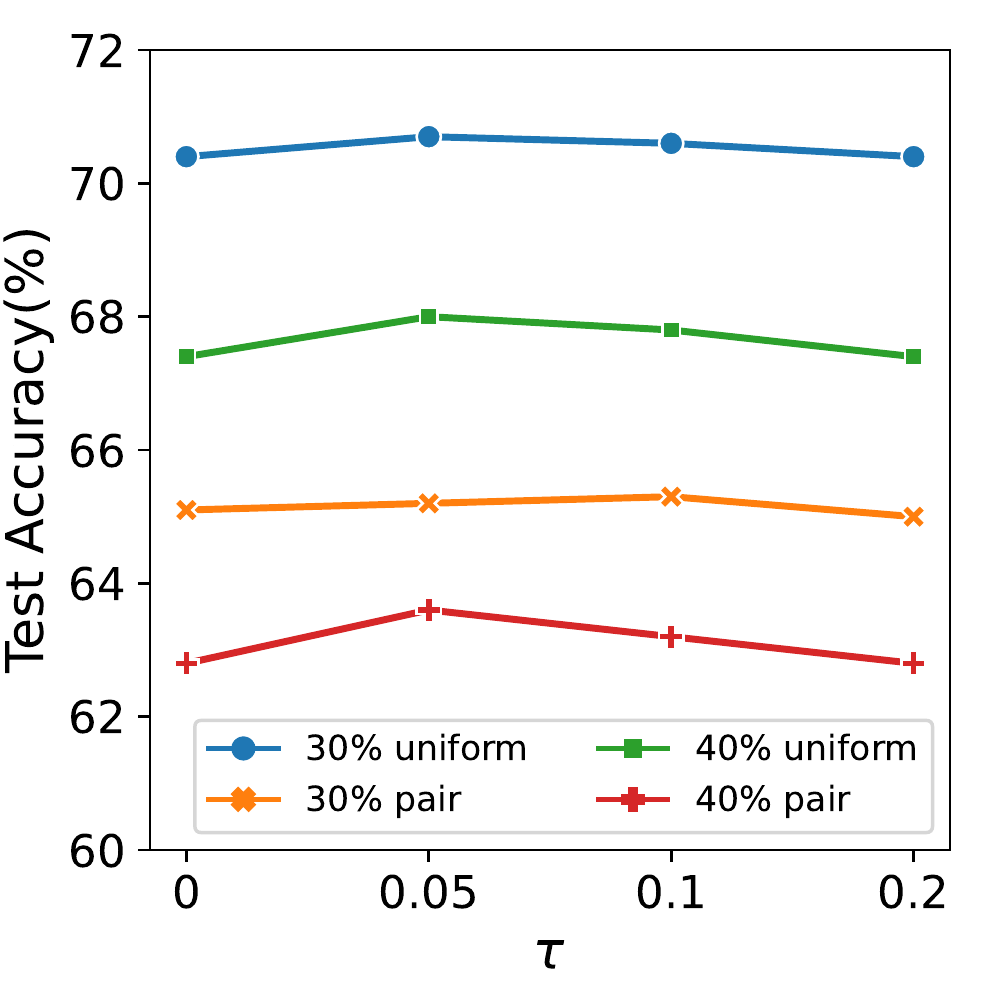}
\label{fig:hypertsmall:b}}
\subfigure[Impact of $\lambda$ on Citeseer]{\includegraphics[width=0.47\linewidth]{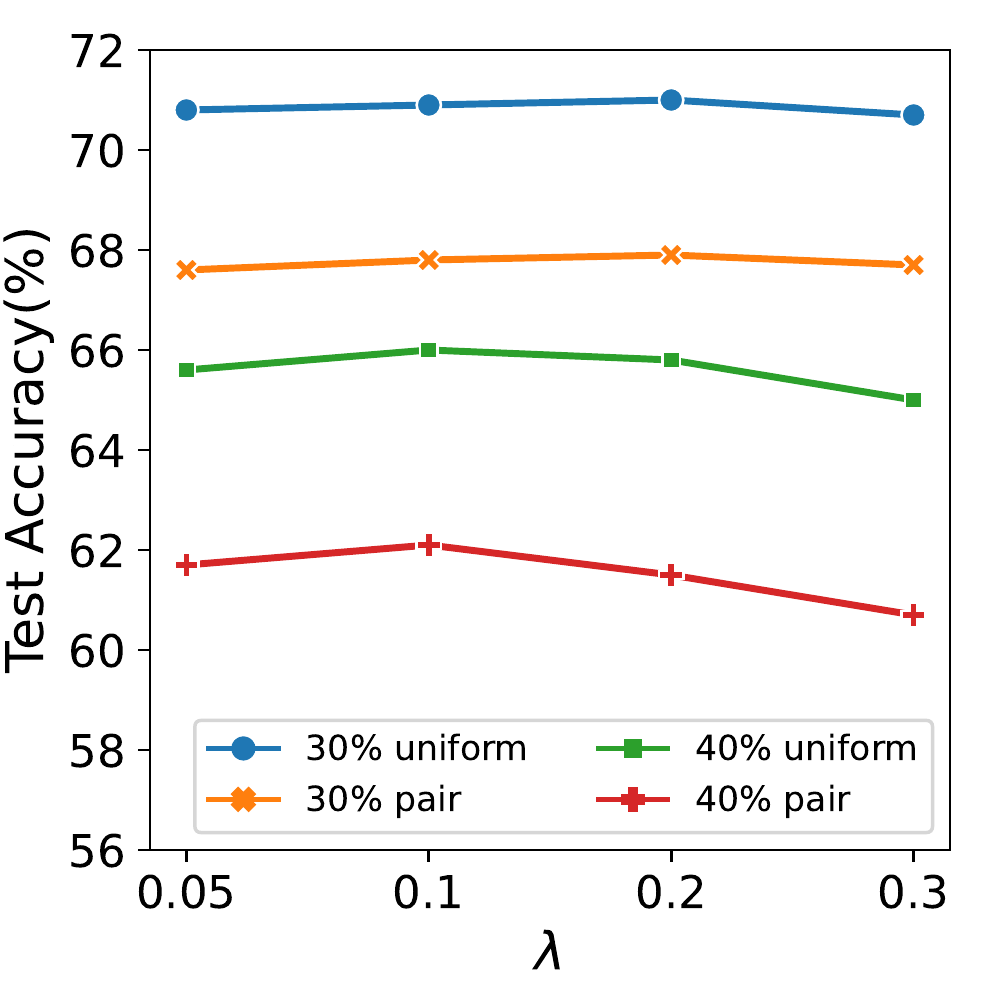}
\label{fig:hyperlambda:a}}
\subfigure[Impact of $\lambda$ on BlogCatalog]{\includegraphics[width=0.47\linewidth]{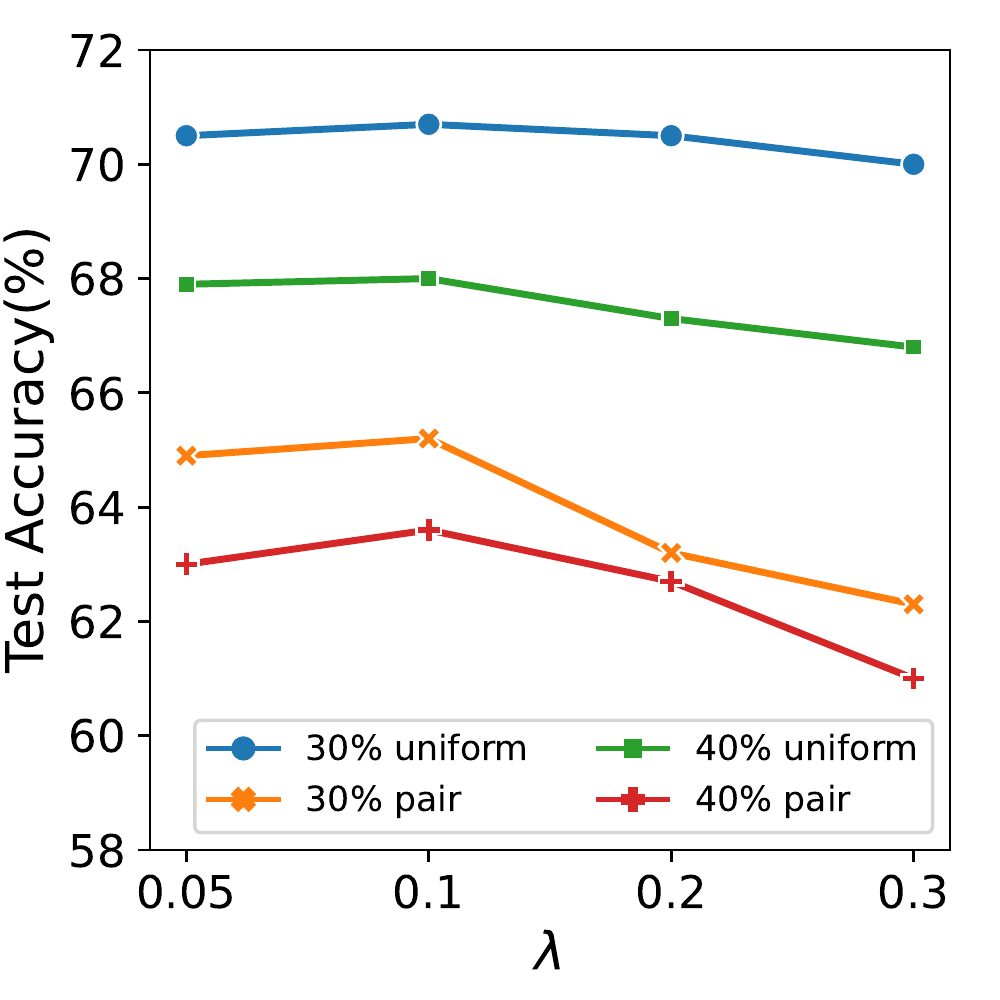}
\label{fig:hyperlambda:b}}
\vspace{-0.4cm}
\caption{Hyper-parameters Sensitivity Analysis.}
\vspace{-0.4cm}
\end{figure}



In this section, we conduct experiments to examine the impact of several important hyper-parameters in our model. The label rate is set as $x$ = 5\% and the noise rate is set as 30\% and 40\% as examples. We report the average results of 5 runs. 

We consider the following key hyper-parameters. (1) In Fig.~\ref{fig:hyperalpha:a} and Fig.~\ref{fig:hyperalpha:b}, we show the impact of the graph reconstruction loss weight $\alpha$. On Citeseer, \model achieves the best results in \textbf{most} cases when $\alpha$ = 0.03.  On BlogCatalog, \model achieves the best results when $\alpha$ = 1 for all the cases.
(2) In Fig.~\ref{fig:hypertsmall:a} and Fig.~\ref{fig:hypertsmall:b}, we show the impact of the edge threshold $\tau$. When $\tau$ = 0, noisy edges may be introduced, and when $\tau$ becomes too large, some potential links would not be identified. On Citeseer, \model achieves the best results when $\tau$ = 0.05 for all the cases. On BlogCatalog, \model achieves the best results in \textbf{most}  cases when $\tau$ = 0.05. (3) In Fig.~\ref{fig:hyperlambda:a} and Fig.~\ref{fig:hyperlambda:b} , we show the impact of the consistency regularization weight $\lambda$. 
\model achieves the best result when $\lambda$ = 0.2 on Citeseer under 30\% uniform and pair noise and $\lambda$ = 0.1 on Citeseer under 40\% uniform and pair noise.  In general, $\lambda$ = 0.1 gives good performances for various scenarios on Citeseer. \model achieves the best results when $\lambda$ = 0.1 on BlogCatalog for all cases.

%% file: files/6.conclusion.tex
\section{Conclusions}
In this paper, we investigated how to train a robust GNN classifier with scarce and noisy node labels. We proposed a novel RTGNN framework which performs explicit noise governance with supplemental supervision. 
Specifically, we classified labeled nodes into clean and noisy ones and adopted reinforcement supervision to correct noisy labels. We also created pseudo labels to provide extra training signals. Moreover, we leveraged consistency regularization to prevent overfitting to noise.
By doing so, \model could enable sufficient learning from clean labels while limiting the impact of noisy ones.
Comprehensive experiments on three widely-used datasets showed superior performance of \model under two types of noise and varying noise rates. We expect that our work would inspire more studies on robust semi-supervised learning on graphs. 



%% file: main.bbl

\begin{thebibliography}{46}


\ifx \showCODEN    \undefined \def \showCODEN     #1{\unskip}     \fi
\ifx \showDOI      \undefined \def \showDOI       #1{#1}\fi
\ifx \showISBNx    \undefined \def \showISBNx     #1{\unskip}     \fi
\ifx \showISBNxiii \undefined \def \showISBNxiii  #1{\unskip}     \fi
\ifx \showISSN     \undefined \def \showISSN      #1{\unskip}     \fi
\ifx \showLCCN     \undefined \def \showLCCN      #1{\unskip}     \fi
\ifx \shownote     \undefined \def \shownote      #1{#1}          \fi
\ifx \showarticletitle \undefined \def \showarticletitle #1{#1}   \fi
\ifx \showURL      \undefined \def \showURL       {\relax}        \fi
\providecommand\bibfield[2]{#2}
\providecommand\bibinfo[2]{#2}
\providecommand\natexlab[1]{#1}
\providecommand\showeprint[2][]{arXiv:#2}

\bibitem[Alam et~al\mbox{.}(2018)]%
        {alam2018graph}
\bibfield{author}{\bibinfo{person}{Firoj Alam}, \bibinfo{person}{Shafiq Joty},
  {and} \bibinfo{person}{Muhammad Imran}.} \bibinfo{year}{2018}\natexlab{}.
\newblock \showarticletitle{Graph based semi-supervised learning with
  convolution neural networks to classify crisis related tweets}. In
  \bibinfo{booktitle}{\emph{12th International AAAI Conference on Web and
  Social Media}}.
\newblock


\bibitem[Arpit et~al\mbox{.}(2017)]%
        {arpit2017closer}
\bibfield{author}{\bibinfo{person}{Devansh Arpit},
  \bibinfo{person}{Stanis{\l}aw Jastrz{\k{e}}bski}, \bibinfo{person}{Nicolas
  Ballas}, \bibinfo{person}{David Krueger}, \bibinfo{person}{Emmanuel Bengio},
  \bibinfo{person}{Maxinder~S Kanwal}, \bibinfo{person}{Tegan Maharaj},
  \bibinfo{person}{Asja Fischer}, \bibinfo{person}{Aaron Courville},
  \bibinfo{person}{Yoshua Bengio}, {et~al\mbox{.}}}
  \bibinfo{year}{2017}\natexlab{}.
\newblock \showarticletitle{A closer look at memorization in deep networks}. In
  \bibinfo{booktitle}{\emph{International Conference on Machine Learning}}.
  PMLR, \bibinfo{pages}{233--242}.
\newblock


\bibitem[Bruna et~al\mbox{.}(2014)]%
        {bruna13spectral}
\bibfield{author}{\bibinfo{person}{Joan Bruna}, \bibinfo{person}{Wojciech
  Zaremba}, \bibinfo{person}{Arthur Szlam}, {and} \bibinfo{person}{Yann
  LeCun}.} \bibinfo{year}{2014}\natexlab{}.
\newblock \showarticletitle{Spectral Networks and Locally Connected Networks on
  Graphs}. In \bibinfo{booktitle}{\emph{2nd International Conference on
  Learning Representations, {ICLR}}}.
\newblock


\bibitem[Chen et~al\mbox{.}(2019)]%
        {chen2019deep}
\bibfield{author}{\bibinfo{person}{Yu Chen}, \bibinfo{person}{Lingfei Wu},
  {and} \bibinfo{person}{Mohammed~J Zaki}.} \bibinfo{year}{2019}\natexlab{}.
\newblock \showarticletitle{Deep iterative and adaptive learning for graph
  neural networks}.
\newblock \bibinfo{journal}{\emph{arXiv preprint arXiv:1912.07832}}
  (\bibinfo{year}{2019}).
\newblock


\bibitem[Cosmo et~al\mbox{.}(2020)]%
        {cosmo2020latent}
\bibfield{author}{\bibinfo{person}{Luca Cosmo}, \bibinfo{person}{Anees Kazi},
  \bibinfo{person}{Seyed-Ahmad Ahmadi}, \bibinfo{person}{Nassir Navab}, {and}
  \bibinfo{person}{Michael Bronstein}.} \bibinfo{year}{2020}\natexlab{}.
\newblock \showarticletitle{Latent-graph learning for disease prediction}. In
  \bibinfo{booktitle}{\emph{International Conference on Medical Image Computing
  and Computer-Assisted Intervention}}. Springer, \bibinfo{pages}{643--653}.
\newblock


\bibitem[Dai et~al\mbox{.}(2021)]%
        {dai2021nrgnn}
\bibfield{author}{\bibinfo{person}{Enyan Dai}, \bibinfo{person}{Charu
  Aggarwal}, {and} \bibinfo{person}{Suhang Wang}.}
  \bibinfo{year}{2021}\natexlab{}.
\newblock \showarticletitle{{NRGNN}: Learning a Label Noise Resistant Graph
  Neural Network on Sparsely and Noisily Labeled Graphs}. In
  \bibinfo{booktitle}{\emph{Proceedings of the 27th ACM SIGKDD Conference on
  Knowledge Discovery \& Data Mining}}. \bibinfo{pages}{227--236}.
\newblock


\bibitem[Dai et~al\mbox{.}(2022)]%
        {dai2022towards}
\bibfield{author}{\bibinfo{person}{Enyan Dai}, \bibinfo{person}{Wei Jin},
  \bibinfo{person}{Hui Liu}, {and} \bibinfo{person}{Suhang Wang}.}
  \bibinfo{year}{2022}\natexlab{}.
\newblock \showarticletitle{Towards Robust Graph Neural Networks for Noisy
  Graphs with Sparse Labels}. In \bibinfo{booktitle}{\emph{{WSDM}'22: The 15th
  {ACM} International Conference on Web Search and Data Mining}}.
\newblock


\bibitem[Defferrard et~al\mbox{.}(2016)]%
        {defferrard2016convolutional}
\bibfield{author}{\bibinfo{person}{Micha{\"e}l Defferrard},
  \bibinfo{person}{Xavier Bresson}, {and} \bibinfo{person}{Pierre
  Vandergheynst}.} \bibinfo{year}{2016}\natexlab{}.
\newblock \showarticletitle{Convolutional neural networks on graphs with fast
  localized spectral filtering}.
\newblock \bibinfo{journal}{\emph{Advances in Neural Information Processing
  Systems}}  \bibinfo{volume}{29} (\bibinfo{year}{2016}).
\newblock


\bibitem[Fatemi et~al\mbox{.}(2021)]%
        {fatemi2021slaps}
\bibfield{author}{\bibinfo{person}{Bahare Fatemi}, \bibinfo{person}{Layla
  El~Asri}, {and} \bibinfo{person}{Seyed~Mehran Kazemi}.}
  \bibinfo{year}{2021}\natexlab{}.
\newblock \showarticletitle{{SLAPS}: Self-Supervision Improves Structure
  Learning for Graph Neural Networks}.
\newblock \bibinfo{journal}{\emph{Advances in Neural Information Processing
  Systems}}  \bibinfo{volume}{34} (\bibinfo{year}{2021}).
\newblock


\bibitem[Ghosh et~al\mbox{.}(2017)]%
        {ghosh2017robust}
\bibfield{author}{\bibinfo{person}{Aritra Ghosh}, \bibinfo{person}{Himanshu
  Kumar}, {and} \bibinfo{person}{PS Sastry}.} \bibinfo{year}{2017}\natexlab{}.
\newblock \showarticletitle{Robust loss functions under label noise for deep
  neural networks}. In \bibinfo{booktitle}{\emph{Proceedings of the AAAI
  Conference on Artificial Intelligence}}, Vol.~\bibinfo{volume}{31}.
\newblock


\bibitem[Goldberger and Ben{-}Reuven(2017)]%
        {DBLP:conf/iclr/GoldbergerB17}
\bibfield{author}{\bibinfo{person}{Jacob Goldberger} {and}
  \bibinfo{person}{Ehud Ben{-}Reuven}.} \bibinfo{year}{2017}\natexlab{}.
\newblock \showarticletitle{Training deep neural-networks using a noise
  adaptation layer}. In \bibinfo{booktitle}{\emph{5th International Conference
  on Learning Representations (ICLR) 2017}}.
\newblock


\bibitem[Halcrow et~al\mbox{.}(2020)]%
        {halcrow2020grale}
\bibfield{author}{\bibinfo{person}{Jonathan Halcrow},
  \bibinfo{person}{Alexandru Mosoi}, \bibinfo{person}{Sam Ruth}, {and}
  \bibinfo{person}{Bryan Perozzi}.} \bibinfo{year}{2020}\natexlab{}.
\newblock \showarticletitle{Grale: Designing networks for graph learning}. In
  \bibinfo{booktitle}{\emph{Proceedings of the 26th ACM SIGKDD International
  Conference on Knowledge Discovery \& Data Mining}}.
  \bibinfo{pages}{2523--2532}.
\newblock


\bibitem[Hamilton et~al\mbox{.}(2017)]%
        {hamilton2017inductive}
\bibfield{author}{\bibinfo{person}{Will Hamilton}, \bibinfo{person}{Zhitao
  Ying}, {and} \bibinfo{person}{Jure Leskovec}.}
  \bibinfo{year}{2017}\natexlab{}.
\newblock \showarticletitle{Inductive representation learning on large graphs}.
\newblock \bibinfo{journal}{\emph{Advances in Neural Information Processing
  Systems}}  \bibinfo{volume}{30} (\bibinfo{year}{2017}).
\newblock


\bibitem[Han et~al\mbox{.}(2018)]%
        {han2018co}
\bibfield{author}{\bibinfo{person}{Bo Han}, \bibinfo{person}{Quanming Yao},
  \bibinfo{person}{Xingrui Yu}, \bibinfo{person}{Gang Niu},
  \bibinfo{person}{Miao Xu}, \bibinfo{person}{Weihua Hu}, \bibinfo{person}{Ivor
  Tsang}, {and} \bibinfo{person}{Masashi Sugiyama}.}
  \bibinfo{year}{2018}\natexlab{}.
\newblock \showarticletitle{Co-teaching: Robust training of deep neural
  networks with extremely noisy labels}.
\newblock \bibinfo{journal}{\emph{Advances in Neural Information Processing
  Systems}}  \bibinfo{volume}{31} (\bibinfo{year}{2018}).
\newblock


\bibitem[Han et~al\mbox{.}(2016)]%
        {han15deep}
\bibfield{author}{\bibinfo{person}{Song Han}, \bibinfo{person}{Huizi Mao},
  {and} \bibinfo{person}{William~J. Dally}.} \bibinfo{year}{2016}\natexlab{}.
\newblock \showarticletitle{Deep Compression: Compressing Deep Neural Network
  with Pruning, Trained Quantization and {Huffman} Coding}. In
  \bibinfo{booktitle}{\emph{4th International Conference on Learning
  Representations (ICLR)}}, \bibfield{editor}{\bibinfo{person}{Yoshua Bengio}
  {and} \bibinfo{person}{Yann LeCun}} (Eds.).
\newblock


\bibitem[Huang et~al\mbox{.}(2019)]%
        {huang2019o2u}
\bibfield{author}{\bibinfo{person}{Jinchi Huang}, \bibinfo{person}{Lie Qu},
  \bibinfo{person}{Rongfei Jia}, {and} \bibinfo{person}{Binqiang Zhao}.}
  \bibinfo{year}{2019}\natexlab{}.
\newblock \showarticletitle{{O2U}u-net: A simple noisy label detection approach
  for deep neural networks}. In \bibinfo{booktitle}{\emph{Proceedings of the
  IEEE/CVF International Conference on Computer Vision}}.
  \bibinfo{pages}{3326--3334}.
\newblock


\bibitem[Jiang et~al\mbox{.}(2018)]%
        {jiang2018mentornet}
\bibfield{author}{\bibinfo{person}{Lu Jiang}, \bibinfo{person}{Zhengyuan Zhou},
  \bibinfo{person}{Thomas Leung}, \bibinfo{person}{Li-Jia Li}, {and}
  \bibinfo{person}{Li Fei-Fei}.} \bibinfo{year}{2018}\natexlab{}.
\newblock \showarticletitle{{MentorNet}: Learning data-driven curriculum for
  very deep neural networks on corrupted labels}. In
  \bibinfo{booktitle}{\emph{International Conference on Machine Learning}}.
  PMLR, \bibinfo{pages}{2304--2313}.
\newblock


\bibitem[Jin et~al\mbox{.}(2020)]%
        {jin2020graph}
\bibfield{author}{\bibinfo{person}{Wei Jin}, \bibinfo{person}{Yao Ma},
  \bibinfo{person}{Xiaorui Liu}, \bibinfo{person}{Xianfeng Tang},
  \bibinfo{person}{Suhang Wang}, {and} \bibinfo{person}{Jiliang Tang}.}
  \bibinfo{year}{2020}\natexlab{}.
\newblock \showarticletitle{Graph structure learning for robust graph neural
  networks}. In \bibinfo{booktitle}{\emph{Proceedings of the 26th ACM SIGKDD
  International Conference on Knowledge Discovery \& Data Mining}}.
  \bibinfo{pages}{66--74}.
\newblock


\bibitem[Kazi et~al\mbox{.}(2022)]%
        {kazi2022differentiable}
\bibfield{author}{\bibinfo{person}{Anees Kazi}, \bibinfo{person}{Luca Cosmo},
  \bibinfo{person}{Seyed-Ahmad Ahmadi}, \bibinfo{person}{Nassir Navab}, {and}
  \bibinfo{person}{Michael Bronstein}.} \bibinfo{year}{2022}\natexlab{}.
\newblock \showarticletitle{Differentiable graph module ({DGM}) for graph
  convolutional networks}.
\newblock \bibinfo{journal}{\emph{IEEE Transactions on Pattern Analysis and
  Machine Intelligence}} (\bibinfo{year}{2022}).
\newblock


\bibitem[Kipf and Welling(2016a)]%
        {kipf2016semi}
\bibfield{author}{\bibinfo{person}{Thomas~N Kipf} {and} \bibinfo{person}{Max
  Welling}.} \bibinfo{year}{2016}\natexlab{a}.
\newblock \showarticletitle{Semi-supervised classification with graph
  convolutional networks}.
\newblock \bibinfo{journal}{\emph{arXiv preprint arXiv:1609.02907}}
  (\bibinfo{year}{2016}).
\newblock


\bibitem[Kipf and Welling(2016b)]%
        {kipf2016variational}
\bibfield{author}{\bibinfo{person}{Thomas~N Kipf} {and} \bibinfo{person}{Max
  Welling}.} \bibinfo{year}{2016}\natexlab{b}.
\newblock \showarticletitle{Variational graph auto-encoders}.
\newblock \bibinfo{journal}{\emph{arXiv preprint arXiv:1611.07308}}
  (\bibinfo{year}{2016}).
\newblock


\bibitem[Louizos et~al\mbox{.}(2018)]%
        {louizos18learning}
\bibfield{author}{\bibinfo{person}{Christos Louizos}, \bibinfo{person}{Max
  Welling}, {and} \bibinfo{person}{Diederik~P. Kingma}.}
  \bibinfo{year}{2018}\natexlab{}.
\newblock \showarticletitle{Learning Sparse Neural Networks through $L_0$
  Regularization}. In \bibinfo{booktitle}{\emph{6th International Conference on
  Learning Representations (ICLR)}}.
\newblock


\bibitem[Ma et~al\mbox{.}(2020)]%
        {ma2020normalized}
\bibfield{author}{\bibinfo{person}{Xingjun Ma}, \bibinfo{person}{Hanxun Huang},
  \bibinfo{person}{Yisen Wang}, \bibinfo{person}{Simone Romano},
  \bibinfo{person}{Sarah Erfani}, {and} \bibinfo{person}{James Bailey}.}
  \bibinfo{year}{2020}\natexlab{}.
\newblock \showarticletitle{Normalized loss functions for deep learning with
  noisy labels}. In \bibinfo{booktitle}{\emph{International Conference on
  Machine Learning}}. PMLR, \bibinfo{pages}{6543--6553}.
\newblock


\bibitem[Malach and Shalev-Shwartz(2017)]%
        {malach2017decoupling}
\bibfield{author}{\bibinfo{person}{Eran Malach} {and} \bibinfo{person}{Shai
  Shalev-Shwartz}.} \bibinfo{year}{2017}\natexlab{}.
\newblock \showarticletitle{Decoupling "when to update" from "how to update"}.
\newblock \bibinfo{journal}{\emph{Advances in Neural Information Processing
  Systems}}  \bibinfo{volume}{30} (\bibinfo{year}{2017}).
\newblock


\bibitem[McPherson et~al\mbox{.}(2001)]%
        {mcpherson2001birds}
\bibfield{author}{\bibinfo{person}{Miller McPherson}, \bibinfo{person}{Lynn
  Smith-Lovin}, {and} \bibinfo{person}{James~M Cook}.}
  \bibinfo{year}{2001}\natexlab{}.
\newblock \showarticletitle{Birds of a Feather: Homophily in Social Networks}.
\newblock \bibinfo{journal}{\emph{Annual Review of Sociology}}
  \bibinfo{volume}{27}, \bibinfo{number}{1} (\bibinfo{year}{2001}),
  \bibinfo{pages}{415--444}.
\newblock


\bibitem[Nguyen et~al\mbox{.}(2020)]%
        {nguyen2019self}
\bibfield{author}{\bibinfo{person}{Duc~Tam Nguyen},
  \bibinfo{person}{Chaithanya~Kumar Mummadi},
  \bibinfo{person}{Thi{-}Phuong{-}Nhung Ngo}, \bibinfo{person}{Thi Hoai~Phuong
  Nguyen}, \bibinfo{person}{Laura Beggel}, {and} \bibinfo{person}{Thomas
  Brox}.} \bibinfo{year}{2020}\natexlab{}.
\newblock \showarticletitle{{SELF:} Learning to Filter Noisy Labels with
  Self-Ensembling}. In \bibinfo{booktitle}{\emph{8th International Conference
  on Learning Representations (ICLR)}}.
\newblock


\bibitem[NT et~al\mbox{.}(2019)]%
        {nt2019learning}
\bibfield{author}{\bibinfo{person}{Hoang NT}, \bibinfo{person}{Choong~Jun Jin},
  {and} \bibinfo{person}{Tsuyoshi Murata}.} \bibinfo{year}{2019}\natexlab{}.
\newblock \showarticletitle{Learning graph neural networks with noisy labels}.
\newblock \bibinfo{journal}{\emph{arXiv preprint arXiv:1905.01591}}
  (\bibinfo{year}{2019}).
\newblock


\bibitem[Pan et~al\mbox{.}(2018)]%
        {pan2018adversarially}
\bibfield{author}{\bibinfo{person}{Shirui Pan}, \bibinfo{person}{Ruiqi Hu},
  \bibinfo{person}{Guodong Long}, \bibinfo{person}{Jing Jiang},
  \bibinfo{person}{Lina Yao}, {and} \bibinfo{person}{Chengqi Zhang}.}
  \bibinfo{year}{2018}\natexlab{}.
\newblock \showarticletitle{Adversarially regularized graph autoencoder for
  graph embedding}. In \bibinfo{booktitle}{\emph{Proceedings of the 27th
  International Joint Conference on Artificial Intelligence}}.
  \bibinfo{pages}{2609--2615}.
\newblock


\bibitem[Patrini et~al\mbox{.}(2017)]%
        {patrini2017making}
\bibfield{author}{\bibinfo{person}{Giorgio Patrini},
  \bibinfo{person}{Alessandro Rozza}, \bibinfo{person}{Aditya Krishna~Menon},
  \bibinfo{person}{Richard Nock}, {and} \bibinfo{person}{Lizhen Qu}.}
  \bibinfo{year}{2017}\natexlab{}.
\newblock \showarticletitle{Making deep neural networks robust to label noise:
  A loss correction approach}. In \bibinfo{booktitle}{\emph{Proceedings of the
  IEEE Conference on Computer Vision and Pattern Recognition}}.
  \bibinfo{pages}{1944--1952}.
\newblock


\bibitem[Pleiss et~al\mbox{.}(2020)]%
        {pleiss2020identifying}
\bibfield{author}{\bibinfo{person}{Geoff Pleiss}, \bibinfo{person}{Tianyi
  Zhang}, \bibinfo{person}{Ethan Elenberg}, {and} \bibinfo{person}{Kilian~Q
  Weinberger}.} \bibinfo{year}{2020}\natexlab{}.
\newblock \showarticletitle{Identifying mislabeled data using the area under
  the margin ranking}.
\newblock \bibinfo{journal}{\emph{Advances in Neural Information Processing
  Systems}}  \bibinfo{volume}{33} (\bibinfo{year}{2020}),
  \bibinfo{pages}{17044--17056}.
\newblock


\bibitem[Veli{\v{c}}kovi{\'c} et~al\mbox{.}(2017)]%
        {velivckovic2017graph}
\bibfield{author}{\bibinfo{person}{Petar Veli{\v{c}}kovi{\'c}},
  \bibinfo{person}{Guillem Cucurull}, \bibinfo{person}{Arantxa Casanova},
  \bibinfo{person}{Adriana Romero}, \bibinfo{person}{Pietro Lio}, {and}
  \bibinfo{person}{Yoshua Bengio}.} \bibinfo{year}{2017}\natexlab{}.
\newblock \showarticletitle{Graph attention networks}.
\newblock \bibinfo{journal}{\emph{arXiv preprint arXiv:1710.10903}}
  (\bibinfo{year}{2017}).
\newblock


\bibitem[Wang et~al\mbox{.}(2019a)]%
        {wang2019semi}
\bibfield{author}{\bibinfo{person}{Daixin Wang}, \bibinfo{person}{Jianbin Lin},
  \bibinfo{person}{Peng Cui}, \bibinfo{person}{Quanhui Jia},
  \bibinfo{person}{Zhen Wang}, \bibinfo{person}{Yanming Fang},
  \bibinfo{person}{Quan Yu}, \bibinfo{person}{Jun Zhou},
  \bibinfo{person}{Shuang Yang}, {and} \bibinfo{person}{Yuan Qi}.}
  \bibinfo{year}{2019}\natexlab{a}.
\newblock \showarticletitle{A Semi-supervised Graph Attentive Network for
  Financial Fraud Detection}. In \bibinfo{booktitle}{\emph{2019 IEEE
  International Conference on Data Mining (ICDM)}}. IEEE,
  \bibinfo{pages}{598--607}.
\newblock


\bibitem[Wang et~al\mbox{.}(2021)]%
        {wang2021learning}
\bibfield{author}{\bibinfo{person}{Deng-Bao Wang}, \bibinfo{person}{Yong Wen},
  \bibinfo{person}{Lujia Pan}, {and} \bibinfo{person}{Min-Ling Zhang}.}
  \bibinfo{year}{2021}\natexlab{}.
\newblock \showarticletitle{Learning from noisy labels with complementary loss
  functions}. In \bibinfo{booktitle}{\emph{Proceedings of the AAAI Conference
  on Artificial Intelligence}}, Vol.~\bibinfo{volume}{35}.
  \bibinfo{pages}{10111--10119}.
\newblock


\bibitem[Wang et~al\mbox{.}(2019b)]%
        {wang2019symmetric}
\bibfield{author}{\bibinfo{person}{Yisen Wang}, \bibinfo{person}{Xingjun Ma},
  \bibinfo{person}{Zaiyi Chen}, \bibinfo{person}{Yuan Luo},
  \bibinfo{person}{Jinfeng Yi}, {and} \bibinfo{person}{James Bailey}.}
  \bibinfo{year}{2019}\natexlab{b}.
\newblock \showarticletitle{Symmetric cross entropy for robust learning with
  noisy labels}. In \bibinfo{booktitle}{\emph{Proceedings of the IEEE/CVF
  International Conference on Computer Vision}}. \bibinfo{pages}{322--330}.
\newblock


\bibitem[Wei et~al\mbox{.}(2020)]%
        {wei2020combating}
\bibfield{author}{\bibinfo{person}{Hongxin Wei}, \bibinfo{person}{Lei Feng},
  \bibinfo{person}{Xiangyu Chen}, {and} \bibinfo{person}{Bo An}.}
  \bibinfo{year}{2020}\natexlab{}.
\newblock \showarticletitle{Combating noisy labels by agreement: A joint
  training method with co-regularization}. In
  \bibinfo{booktitle}{\emph{Proceedings of the IEEE/CVF Conference on Computer
  Vision and Pattern Recognition}}. \bibinfo{pages}{13726--13735}.
\newblock


\bibitem[Wu et~al\mbox{.}(2019)]%
        {wu2019net}
\bibfield{author}{\bibinfo{person}{Jun Wu}, \bibinfo{person}{Jingrui He}, {and}
  \bibinfo{person}{Jiejun Xu}.} \bibinfo{year}{2019}\natexlab{}.
\newblock \showarticletitle{{DEMO-Net}: Degree-specific graph neural networks
  for node and graph classification}. In \bibinfo{booktitle}{\emph{Proceedings
  of the 25th ACM SIGKDD International Conference on Knowledge Discovery \&
  Data Mining}}. \bibinfo{pages}{406--415}.
\newblock


\bibitem[Xia et~al\mbox{.}(2021)]%
        {xia2021sample}
\bibfield{author}{\bibinfo{person}{Xiaobo Xia}, \bibinfo{person}{Tongliang
  Liu}, \bibinfo{person}{Bo Han}, \bibinfo{person}{Mingming Gong},
  \bibinfo{person}{Jun Yu}, \bibinfo{person}{Gang Niu}, {and}
  \bibinfo{person}{Masashi Sugiyama}.} \bibinfo{year}{2021}\natexlab{}.
\newblock \showarticletitle{Sample selection with uncertainty of losses for
  learning with noisy labels}.
\newblock \bibinfo{journal}{\emph{arXiv preprint arXiv:2106.00445}}
  (\bibinfo{year}{2021}).
\newblock


\bibitem[Xu et~al\mbox{.}(2018)]%
        {xu2018powerful}
\bibfield{author}{\bibinfo{person}{Keyulu Xu}, \bibinfo{person}{Weihua Hu},
  \bibinfo{person}{Jure Leskovec}, {and} \bibinfo{person}{Stefanie Jegelka}.}
  \bibinfo{year}{2018}\natexlab{}.
\newblock \showarticletitle{How powerful are graph neural networks?}
\newblock \bibinfo{journal}{\emph{arXiv preprint arXiv:1810.00826}}
  (\bibinfo{year}{2018}).
\newblock


\bibitem[Yang et~al\mbox{.}(2016)]%
        {yang2016revisiting}
\bibfield{author}{\bibinfo{person}{Zhilin Yang}, \bibinfo{person}{William
  Cohen}, {and} \bibinfo{person}{Ruslan Salakhudinov}.}
  \bibinfo{year}{2016}\natexlab{}.
\newblock \showarticletitle{Revisiting semi-supervised learning with graph
  embeddings}. In \bibinfo{booktitle}{\emph{International Conference on Machine
  Learning}}. PMLR, \bibinfo{pages}{40--48}.
\newblock


\bibitem[Yu et~al\mbox{.}(2021)]%
        {yu2021divergence}
\bibfield{author}{\bibinfo{person}{Qing Yu}, \bibinfo{person}{Atsushi
  Hashimoto}, {and} \bibinfo{person}{Yoshitaka Ushiku}.}
  \bibinfo{year}{2021}\natexlab{}.
\newblock \showarticletitle{Divergence Optimization for Noisy Universal Domain
  Adaptation}. In \bibinfo{booktitle}{\emph{Proceedings of the IEEE/CVF
  Conference on Computer Vision and Pattern Recognition}}.
  \bibinfo{pages}{2515--2524}.
\newblock


\bibitem[Yu et~al\mbox{.}(2019)]%
        {yu2019does}
\bibfield{author}{\bibinfo{person}{Xingrui Yu}, \bibinfo{person}{Bo Han},
  \bibinfo{person}{Jiangchao Yao}, \bibinfo{person}{Gang Niu},
  \bibinfo{person}{Ivor Tsang}, {and} \bibinfo{person}{Masashi Sugiyama}.}
  \bibinfo{year}{2019}\natexlab{}.
\newblock \showarticletitle{How does disagreement help generalization against
  label corruption?}. In \bibinfo{booktitle}{\emph{International Conference on
  Machine Learning}}. PMLR, \bibinfo{pages}{7164--7173}.
\newblock


\bibitem[Zhang et~al\mbox{.}(2017)]%
        {zhang17understanding}
\bibfield{author}{\bibinfo{person}{Chiyuan Zhang}, \bibinfo{person}{Samy
  Bengio}, \bibinfo{person}{Moritz Hardt}, \bibinfo{person}{Benjamin Recht},
  {and} \bibinfo{person}{Oriol Vinyals}.} \bibinfo{year}{2017}\natexlab{}.
\newblock \showarticletitle{Understanding deep learning requires rethinking
  generalization}. In \bibinfo{booktitle}{\emph{5th International Conference on
  Learning Representations (ICLR)}}.
\newblock


\bibitem[Zhang et~al\mbox{.}(2018)]%
        {zhang2018deep}
\bibfield{author}{\bibinfo{person}{Ying Zhang}, \bibinfo{person}{Tao Xiang},
  \bibinfo{person}{Timothy~M Hospedales}, {and} \bibinfo{person}{Huchuan Lu}.}
  \bibinfo{year}{2018}\natexlab{}.
\newblock \showarticletitle{Deep mutual learning}. In
  \bibinfo{booktitle}{\emph{Proceedings of the IEEE Conference on Computer
  Vision and Pattern Recognition}}. \bibinfo{pages}{4320--4328}.
\newblock


\bibitem[Zhang and Sabuncu(2018)]%
        {zhang2018generalized}
\bibfield{author}{\bibinfo{person}{Zhilu Zhang} {and} \bibinfo{person}{Mert
  Sabuncu}.} \bibinfo{year}{2018}\natexlab{}.
\newblock \showarticletitle{Generalized cross entropy loss for training deep
  neural networks with noisy labels}.
\newblock \bibinfo{journal}{\emph{Advances in Neural Information Processing
  Systems}}  \bibinfo{volume}{31} (\bibinfo{year}{2018}).
\newblock


\bibitem[Zhao et~al\mbox{.}(2021)]%
        {zhao2021data}
\bibfield{author}{\bibinfo{person}{Tong Zhao}, \bibinfo{person}{Yozen Liu},
  \bibinfo{person}{Leonardo Neves}, \bibinfo{person}{Oliver Woodford},
  \bibinfo{person}{Meng Jiang}, {and} \bibinfo{person}{Neil Shah}.}
  \bibinfo{year}{2021}\natexlab{}.
\newblock \showarticletitle{Data augmentation for graph neural networks}. In
  \bibinfo{booktitle}{\emph{Proceedings of the AAAI Conference on Artificial
  Intelligence}}, Vol.~\bibinfo{volume}{35}. \bibinfo{pages}{11015--11023}.
\newblock


\bibitem[Zhou et~al\mbox{.}(2021)]%
        {zhou2021informer}
\bibfield{author}{\bibinfo{person}{Haoyi Zhou}, \bibinfo{person}{Shanghang
  Zhang}, \bibinfo{person}{Jieqi Peng}, \bibinfo{person}{Shuai Zhang},
  \bibinfo{person}{Jianxin Li}, \bibinfo{person}{Hui Xiong}, {and}
  \bibinfo{person}{Wancai Zhang}.} \bibinfo{year}{2021}\natexlab{}.
\newblock \showarticletitle{Informer: Beyond efficient transformer for long
  sequence time-series forecasting}. In \bibinfo{booktitle}{\emph{Proceedings
  of the AAAI Conference on Artificial Intelligence}},
  Vol.~\bibinfo{volume}{35}. \bibinfo{pages}{11106--11115}.
\newblock


\end{thebibliography}
